\pdfoutput=1
\documentclass[11pt]{article}

\usepackage[]{acl}

\usepackage{booktabs}
\usepackage{graphicx}
\usepackage{booktabs}
\usepackage{times}
\usepackage{latexsym}
\usepackage{graphicx}
\usepackage{amsmath}
\usepackage{caption}
\usepackage{placeins} 
\usepackage{stmaryrd}
\usepackage[T1]{fontenc}
\usepackage{microtype}
\usepackage{tikz}
\usepackage{amsthm}
\usepackage{amssymb}

\usepackage{subcaption,booktabs}
\usepackage{times}
\usepackage{latexsym}
\usepackage{booktabs}

\usepackage[T1]{fontenc}
\usepackage[utf8]{inputenc}

\usepackage{microtype}

\usepackage{textcomp}
\usepackage{amssymb}
\usepackage{amsmath}
\usepackage{subcaption} 
\usepackage{xspace}

\usepackage{graphicx}
\usepackage{xcolor}

\newtheorem{theorem}{Theorem}[section]
\newtheorem{lemma}[theorem]{Lemma}

\newcommand{\cready}[1]{}

\usepackage{xr}
\usepackage{subfiles}

\makeatletter
\newcommand*{\addFileDependency}[1]{% argument=file name and extension
  \typeout{(#1)}
  \@addtofilelist{#1}
  \IfFileExists{#1}{}{\typeout{No file #1.}}
}
\makeatother

\title{Reinforcement Learning with Large Action Spaces\\ for Neural Machine Translation}

 \author{ 
  Asaf Yehudai, Leshem Choshen, Lior Fox, Omri Abend\\
  Department of Computer Science\\
  The Hebrew University of Jerusalem\\
  {\tt\small \{first.last\}@mail.huji.ac.il}\\
  
}
\begin{document}
\maketitle

\begin{abstract}
Applying Reinforcement learning (RL) following maximum likelihood estimation (MLE) pre-training is 
%a powerful tool that enables optimization of versatile evaluation functions and is frequently used to increase performance in text generation tasks (TG), including 
a versatile method for enhancing neural machine translation (NMT) performance. However, recent work has argued that the gains produced by RL for NMT are mostly due to promoting tokens that have already received a fairly high probability in pre-training. 
%and to effects that are unrelated to the reward function that is optimized. some doubt has recently been cast as to this method's ability to improve performance in non-trivial cases, i.e., cases where the target action (token) is only assigned a low probability by the pre-trained model. 
We hypothesize that the large action space is a main obstacle to RL's effectiveness in MT, and conduct two sets of experiments that lend support to our hypothesis. First, we find that reducing the size of the vocabulary improves RL's effectiveness. Second, we find that effectively reducing the dimension of the action space without changing the vocabulary also yields notable improvement as evaluated by BLEU, semantic similarity, and
human evaluation. Indeed, by initializing the network's final fully connected layer (that maps the network's internal dimension to the vocabulary dimension), with a layer that generalizes over similar actions, we obtain a substantial improvement in RL performance: 1.5 BLEU points on average.\footnote{https://github.com/AsafYehudai/Reinforcement-Learning-with-Large-Action-Spaces-for-Neural-Machine-Translation}

%In the first method, we reduce the size of the target vocabulary by using a small number of words part. Using this method on our neural machine translation (NMT) experiments, we show an improvement of three and a half BLEU points on average.
%Moreover, we show that using this method improves the pre-trained model, even where it was not close to producing the correct output initially.

%In the second method, we aim to decrease the size of action space by freezing the last fully connected (FC) embedding layer thus implicitly changing the action space size to be the size of the embedding space. Our results show that This method cuts the number of parameters by half without harming performance when freezing the pre-trained embedding. A variant of this approach is to transfer a quality pre-trained embedding as the FC layer. Freezing the embedding layer makes it more important to choose its values carefully. We indeed show that BERT embedding poses desirable attributes for RL training that can enhance generalization across similar actions. Applying the freezing method with BERT embedding not only increases performance but also helps the model improve where it was not close to being correct in pre-trained.
\end{abstract}

\section{Introduction} \label{sec:Intro} 

%\oa{commeneted out parag here from the abstract; it's more fitting to the intro; you can salvage it, Asaf: related to lines 102-112, until bert part}
%supports effective learning and improves BLEU by 3.5 points on average. Second, freezing the last layer. Simulations show that if similar actions share embeddings, training is improved. Freezing the last layer, further improves it. On full scale MT, freezing cuts the number of parameters by half, while preserving performance. We show BERT embeddings relates synonyms better than MT trained embeddings. Indeed, loading BERT's last layer parameters increases both time and performance. Overall, we deduce the large action space is a severe problem, which can be alleviated by decreasing the vocabulary size, starting RL from well trained embeddings and freezing the last layer.

%Machine translation (MT) is a sub-field of natural language processing (NLP) that investigates how to translate text or speech from one language to another using software. Recently, with the development of Deep Learning, a new kind of method based on Artificial neural networks named Neural Machine Translation (NMT) \citep{bahdanau2014neural, hassan2018achieving, wu2018study} become more and more popular. NMT surpassed previous methods performance without the demand for heavily hand-crafted engineering efforts.

The standard training method for sequence-to-sequence tasks, specifically for NMT is to maximize the likelihood of a token in the target sentence, given a gold standard prefix (henceforth, maximum likelihood estimation or MLE). However, despite the strong performance displayed by MLE-trained models, this token-level objective function is limited in its ability to penalize sequence-level errors and is at odds with the sequence-level evaluation metrics it aims to improve. 
One appealing method for addressing this gap is applying policy gradient methods that allow incorporating non-differentiable reward functions, such as the ones often used for MT evaluation  \citep[][see \S\ref{sec:Related_Work}]{shen-etal-2016-minimum}.
For brevity, we will refer to these methods simply as RL.

%\oa{add ref}\lc{of an evaluation? of something that incorporates?}\oa{of works on RL for NMT}

The RL training procedure consists of several steps: (1) generating a translation with the pre-trained MLE model, (2) computing some sequence-level reward function, usually, one that assesses the similarity of the generated translation and a reference, and (3) updating the model so that its future outputs receive higher rewards. The method's flexibility, as well as its ability to address the exposure bias \citep{Ranzato2016SequenceLT, wang2020exposure},
%\oa{ref, you can also cite Wang et al. from ACL 2020 here}\lc{which one is that? there are 120 papers with the name wang on them in acl2020}\oa{I meant \url{https://arxiv.org/abs/2005.03642}}
makes RL an appealing avenue for improving NMT performance.
However, a recent study 
\citep[C19;][]{choshen2019weaknesses}
suggests that current RL practices are likely to improve the prediction of target tokens only where the MLE model has already assigned that token a fairly high probability. 
%C19 further suggest that observed gains may be due to changes in the shape of the distribution curve, independently of the training signal. 

In this work, we observe that one main difference between NMT and other tasks in which RL methods excel is the size of the action space. Typically, the size of the action space in NMT includes all tokens in the vocabulary, usually tens of thousands. By contrast, common RL settings have either small discrete action spaces (e.g., Atari games \citep{Mnih2013PlayingAW}), or continuous action spaces of low dimension (e.g., MuJoCo \citep{todorov2012mujoco} and similar control problems). Intuitively, RL takes (samples) actions and assesses their outcome, unlike supervised learning (MLE) which directly receives a score for all actions. Therefore, a large action space will make RL less efficient, as individual actions have to be sampled in order to assess their quality.
%Therefore, the number of actions that RL assesses grows with the size of the action space.
Accordingly, we experiment with two methods for decreasing the size of the action space and evaluate their impact on RL's effectiveness.
 
%\oa{you need to start from a more gentle introduction, talk about NMT, how it's usually trained, how does RL fit in, then dive into the details (keeping it high level of course) A: is it better? new comments on the new paragraphs?}

We begin by decreasing the vocabulary size (or equivalently, the number of actions), conducting experiments in low-resource settings on translating four languages into English, using BLEU both as the reward function and the evaluation metric. Our results show that RL yields a considerably larger performance increase (about $1$ BLEU point on average) over MLE training than is achieved by RL with the standard vocabulary size.
%Furthermore, MLE training gain $\sim 10$ BLEU points more as consequence of this modification.
% To ensure a level ground, we use pre-trained models with similar performance in both settings. 
Moreover, our findings indicate that reducing the size of the vocabulary can improve upon the MLE model even in cases where it was not close to being correct. See \S\ref{sec:first}.
%(see \S\ref{sec:Method} and \S\ref{sec:Experiments}).

However, in some cases, it may be undesirable or unfeasible to change the vocabulary. We therefore experiment with two methods that effectively, reduce the dimensionality of the action space without changing the vocabulary. We note that generally in NMT architectures, the dimensionality of the decoder's internal layers (henceforth, $d$) is significantly smaller than the target vocabulary size (henceforth, $|V_T|$), which is the size of the action space. A fully connected layer is generally used to map the internal representation to suitable outputs. We may therefore refer to the rows of the matrix (parameters) of this layer, as {\it target embeddings}, mapping the network's internal low-dimensional representation back to the vocabulary size, the actions.
We use this term to underscore the analogy between 
the network's first embedding layer, mapping vectors of dimension $|V_T|$ to vectors of dimension $d$, and target embeddings that work in an inverse fashion. 
Indeed, it is often the case \citep[e.g., in BERT,][]{Devlin_2019} that the weights of the source and target embeddings are shared during training, emphasizing the relation between the two.

%If source embeddings are the lowest layer of the network, transforming a token to a representation, target embedding is the weights of the last layer, transforming a representation to token logits. 
Using this terminology, we show in simulations (\S\ref{sec:simulation}) that when similar actions share target embeddings, RL is more effective. Moreover, when target embeddings are initialized based on high-quality embeddings (BERT's in our case), freezing them during RL yields further improvement still.
We obtain similar results when experimenting on NMT. 
Indeed, using BERT's embeddings for target embeddings improves performance on the four language pairs, and freezing them yields an additional improvement on both MLE and RL as reported by both automatic metrics and human evaluation.
Both initialization and freezing are novel in the context of RL training for NMT.
Moreover, when using BERT's embeddings, RL's ability to improve performance on target tokens to which the pre-trained MLE model did not assign a high probability, is enhanced 
(\S\ref{sec:NMT_EXP}).

\section{Background} \label{sec:Related_Work} 

\subsection{RL in Machine Translation}\label{sec:back_rl_nmt}

% \oa{aren't these two following paragraphs repetitious with the intro? distill from here the new details and discard the rest}
% Neural machine translation systems are commonly optimized using maximum likelihood estimation (MLE). Such training maximizes the single word conditional distribution (given a gold standard prefix), rather than the probability of the entire sentence. 

RL is used in text generation (TG) for its ability to incorporate non-differentiable signals, to tackle the exposure bias, and to introduce sequence-level constraints.
The latter two are persistent challenges in the development of TG systems, and have also been addressed by non-RL methods \citep[e.g.,][]{Zhang2019BridgingTG,Ren2019UnsupervisedNM}. 
In addition, RL is grounded within a broad theoretical and empirical literature, which adds to its appeal.

% RL-based NMT methods generally operate by generating a translation (from the distribution defined by the network), comparing it to the reference, receiving some reward based on their similarity, and finally updating model parameters to increase future rewards using policy gradient methods. This method allows training systems to optimize non-differentiable score functions (such as BLEU), common in MT evaluation.

%These properties have led to much interest in RL for TG in general \citep{Shah2018BootstrappingAN} and NMT in particular \citep{Wu2018ASO}. Numerous policy gradient methods are commonly used, 
%notably REINFORCE \citep{Williams92simplestatistical}, and variants thereof \citep[e.g.,][]{ranzato2016sequence, Edunov_2018}, actor-critic methods \citep{bahdanau2017actorcritic} and Minimum Risk Training \citep[MRT; e.g.,][]{och2003minimum, shen-etal-2016-minimum}. Another common use
%of RL is for training GANs \citep{yang2017improving, wu2018adversarial, tevet2019evaluating, Caccia2020Language}. However, despite increasing interest and strong results, only a handful of works studied the source of observed performance gains by RL in NLP and its training dynamics, and some of these have suggested that RL's gains are partly due to  artifacts \citep{caccia2018language,choshen2019weaknesses}.

These properties have led to much interest in RL for TG in general \citep{Shah2018BootstrappingAN} and NMT in particular \citep{Wu2018ASO}. Numerous policy gradient methods are commonly used, 
notably REINFORCE \citep{Williams92simplestatistical}, and Minimum Risk Training \citep[MRT; e.g.,][]{och2003minimum, shen-etal-2016-minimum}. However, despite increasing interest and strong results, only a handful of works studied the source of observed performance gains by RL in NLP and its training dynamics, and some of these have suggested that RL's gains are partly due to artifacts \citep{caccia2018language,choshen2019weaknesses}.

In a recent paper, C19 showed that existing RL training protocols for MT (REINFORCE and MRT) take a prohibitively long time to converge. Their results suggest that RL practices in MT are likely to improve performance only where the MLE parameters are already close to yielding the correct translation. They further suggest that observed gains may be due to effects unrelated to the training signal, but rather from changes in the shape of the distribution curve. These results may suggest that one of the drawbacks of RL is the uncommonly large action space, which in TG includes all tokens in the vocabulary, typically tens of thousands of actions or more.

To the best of our knowledge, no previous work considered the challenge of large action spaces in TG, and relatively few studies considered it in different contexts.
One line of work assumed prior domain knowledge about the problem, and partitioned actions into sub-groups \citep{sharma2017learning}, or similar to our approach, embedding actions in a continuous space where some metric over this space allows generalization over similar actions \citep{dulacarnold2016deep}. More recent work proposed to learn target embeddings when the underlying structure of the action space is apriori unknown using expert demonstrations \citep{Tennenholtz2019TheNL,Chandak2019LearningAR}.
%or to simultaneously train an abstract policy that acts in some latent action space together with a mapping function from the latent space to the concrete action space of the problem \citep{Chandak2019LearningAR}.  

This paper establishes that the large action spaces are a limiting factor in the application of RL for NMT, and propose methods to tackle this challenge. Our techniques restrict the size of the embedding space, either explicitly or implicitly by using an underlying continuous representation. 

%Little previous work addressed problems stemming from large action spaces, and to our knowledge no previous work has done so for text generation.
%\citet{sharma2017learning} showed that given outside information about the actions, better performance can be achieved on Atari 2600 games when actions are factored into their primary categories. To deal with discrete actions that might have an underlying continuous representation, \citet{van2009using} used policy gradients with continuous actions and selected the nearest discrete action. This work was extended by \citet{dulacarnold2016deep} for larger domains\oa{what are larger domains?}, where they performed an action representation lookup\oa{what is that? try to describe it in more high level}.\lc{I wanted to just combine the two, but then I wondered, how could the 2016 paper extend the 2019 paper? Asaf: the first one is from 2009} However, They do not offer a method in which such target embedding can be found. Recent work also showed how action representations can be learned using data from expert demonstrations \citep{Tennenholtz2019TheNL}, or how embedding is used as part of the policy's structure in order to train an agent even without prior knowledge \citep{Chandak2019LearningAR}.

%In this work, we address those issues and suggest techniques to tackle the large action space problem in RL for NMT. Our techniques aim to restrict the size of the embedding space by using an underlying continuous representation. 

%%%%%%%%%%%%%%%%%%%%%%%%%%%%%%%%%%%%%%%%%%%%%%
\subsection{Technical Background and Notation} \label{sec:Technical_Background} 

%In this section, we describe the attention-based sequence-to-sequence learning framework for NMT, and elaborate on applying RL in training.

\paragraph{Notation.}
We denote the source sentence with $X = (x_1, ..., x_S)$ and the reference sentence with $Y = (y_1, ..., y_{T})$.
Given $X$, the network generates a sentence in the target language $Y' = (y'_1, ..., y'_M)$. Target tokens are taken from a  vocabulary $V_{T}$. During inference, at each step $i$, the probability of generating a token $y'_i \in V_{T}$ is conditioned on the sentence and the predicted tokens, i.e.,  $P_{\theta}(y'_i|X, y'_{<i})$, where $\theta$ is the model parameters. We assume there is exactly one valid target token, the reference token, as in practice, training is done against a single reference \citep{schulz2018stochastic}\cready{even if not ideal \citep{choshen2019inherent}}. 

% Given $N$ training sentence pairs $\{X^{(j)},Y^{(j)}\}_{j=1}^N$, MLE training of $\theta$ is the maximization of the conditional log likelihood:
% \begin{equation}\label{MLE_eqn}
% \begin{split}
% 	L_{mle} & = \Sigma_{j=1}^N log P_{\theta}(Y^j|X^j) \\
% 	& = \Sigma_{j=1}^N \Sigma_{i=1}^{M_j} log P_{\theta}(y^j_i|y^j_{<i}X^j)
% \end{split}
% \end{equation}
% where $M_j$ is the length of $Y^j$.   

%\oa{what about regularization?}
%From all the encoder-decoder models, Transformer \citep{vaswani2017attention} architecture achieves the best translation quality so far. The most significant difference between the Transformer and previous models architecture is the self-attention mechanism \citep{lin2017structured} that Transformer architecture incorporates in order to compute representations of source and target side sentences. As we will further explain in \S\ref{sec:Experiments_Preliminaries} we used Transformer for our experiments.

\paragraph{NMT with RL.}
%As mentioned, reinforcement learning (RL) can optimize the parameters to explicitly maximize a sentence level metric (e.g., BLEU, METEOR) by incorporating such score function in the objective function.
In RL terminology, one can think of an NMT model as an agent, which interacts with the environment. In this case, the environment state consists of the previous words $y'_{<i}$ and the source sentence $X$.
At each step, the agent selects an action according to its policy, where actions are tokens. The policy is defined by the parameters of the model, i.e., the conditional probability $P_{\theta}(y'_i|y'_{<i}, X)$. Reward is given only once the agent
generates a complete sequence $Y'$.
The standard reward for MT is the sentence level BLEU metric \citep{papineni2002bleu}, matching the evaluation metric. %, denoted by $R(y,y')$, which is defined
%by comparing different n-gram counts from the generated sentence $y' $ with the reference sentence y.\oa{you don't need to explain what BLEU is for an NLP crowd}
% The goal of the RL training is to maximize the expected reward:
% \begin{equation}
% \label{RL_eqn}
% \begin{split}
% 	L_{rl} &= \Sigma_{j=1}^N E_{y' \sim P_\theta(y'|x^i)} R(y,y') \\
% 	&= \Sigma_{j=1}^N \Sigma_{y' \in Y_{All}}  P_\theta(y'|x^i) R(y,y')
% \end{split}
% \end{equation}
% equivalently 
Our goal is to find the parameters that will maximize the expected reward.

% \begin{equation}\label{RL_theta_eqn}
% \begin{split}
% 	\theta^* &=  \argmax_{\theta} \Sigma_{j=1}^N E_{y' \sim P_\theta(y'|x^i)} R(y,y') \\
% 	&=   \argmax_{\theta} \Sigma_{j=1}^N \Sigma_{y' \in Y_All}  P_\theta(y'|x^i) R(y,y')
% \end{split}
% \end{equation}

% Where $Y_{All}$ is the space of all possible translation sentences, which grows exponentially  with the size of the vocabulary. It is therefore unfeasible to maximize the parameters explicitly, motivating the use of policy gradient methods like REINFORCE \citep{Williams92simplestatistical}, which provide an unbiased estimator to the risk's gradient.

%above expectation via sampling, for each source sentence, k (k is a hyperparameter) candidate translation sentences,$\{y'_i\}_{i=1}^k$ , from the policy $P_{\theta}(y|x)$, leading to the objective as maximizing:
% \begin{equation}\label{REINFORCE_eqn_reward}
% 	L_{rl} =  \Sigma_{j=1}^N \Sigma_{s=1}^k R(y,y'_s) \quad where \, \forall i \in [N] \,\, y'_s \sim P_\theta(y'_s|x^i)
% \end{equation}
% equivalently, at each step we updates $\theta$ according to this rule:
% \begin{equation}\label{REINFORCE_eqn_update}
% 	\Delta\theta \propto \frac{1}{k}  \Sigma_{s=1}^k R(y,y'_s) \nabla log( P_\theta(y'_s|x^i))
% \end{equation}

% Another estimation method called 
In this work, we use MRT \citep{och2003minimum,shen2015minimum}, a policy gradient method adapted to MT.
The key idea of this method is to optimize at each step a re-normalized risk, defined only over the sampled batch. Concretely, the expected risk is defined as:
\begin{equation}\label{RISK_eqn}
	L_{risk} = \sum_{u \in U(X)} R(Y,u) \frac{P(u|X)^\beta}{\sum_{u' \in U(X)} P(u'|X)^\beta} 
\end{equation}

where $u$ is a candidate hypothesis sentence,  $U(x)$ is the sample of $k$ candidate hypotheses, $Y$ is the reference, $P$ is the conditional probability that the model assigns a candidate hypothesis $u$ given source sentence $X$, $\beta$ a
smoothness parameter and $R$ is BLEU.
%$P(u|X)$ is the product of the conditional probabilities of the words  in $u$, normalized by its length $n(u)$:

%\vspace{-.5cm}
%\begin{small}
% \begin{equation}\label{con_prob_eq}
% 	P(u|X) = \left(\prod_{j=1}^{n(u)}{P(u_j|X,u_{<j})}\right) ^{n(u)^{-1}}
% \end{equation}
% \end{small}

%We elaborate on hyperparameters and technicalities regarding our use in \S\ref{sec:Experiments_Preliminaries}. 

\section{Methodology} 

\label{sec:Experiments_Preliminaries}

% \paragraph{Architecture.}
% \label{subsec:Arch.}
% We use a similar setup as used by \citet{wieting2019beyond}, adapting their fairSeq-based \citep{ott2019fairseq} codebase to our purposes.\footnote{\url{https://github.com/jwieting/beyond-bleu}} Similar to their Transformer architecture we use gated convolutional encoders and decoders \citep{gehring2017convolutional}.
%\oa{is this standard? I don't know what it is}\lc{It is not, but it is what the others used} 
%We use 4 layers for the encoder and 3 for the decoder, the size of the hidden state is 768 for all layers, and the filter width of the kernels is 3.
\paragraph{Architecture.}
\label{subsec:Arch.}
We use a similar setup as used by \citet{wieting2019beyond}, adapting their fairSeq-based \citep{ott2019fairseq} codebase to our purposes.\footnote{\url{https://github.com/jwieting/beyond-bleu}} Similar to their Transformer architecture we use gated convolutional encoders and decoders \citep{gehring2017convolutional}. We use 4 layers for the encoder and 3 for the decoder, the size of the hidden state is 768 for all layers, and the filter width of the kernels is 3. Additionally, the dimension of the BPE embeddings is set to 768.

\paragraph{Data Prepossessing.} 
We use BPE \citep{sennrich-etal-2016-neural} for tokenization.
The vocabulary size is set to 40K for the combined source and target vocabulary as done by \citet{wieting2019beyond}. For the small target vocabulary experiments, we change the target vocabulary size to 1K and keep the source vocabulary unchanged.

\paragraph{Objective Functions.}

Following \citet{Edunov_2018}, we train models with MLE with label-smoothing %($\ell_{MLE}$) 
\citep{Szegedy_2016, Pereyra2017RegularizingNN} of size 0.1.
For RL, we fine-tune the model with a weighted average of the MRT $L_{risk}$ and the token level loss $L_{mle}$.

Our fine-tuning objective thus becomes:
\begin{equation}\label{fine-tuning objective}
	L_{Average} = \alpha \cdot L_{mle} + (1 -\alpha) \cdot L_{risk}
\end{equation}

We set $\alpha$ to be 0.3 shown to work best by \citet{wu2018study}.
%and our tuning.
We set $\beta$ to 1.
%We set the reward $1 -BLEU(t, h)$ where t is the target and $h$ is the generated hypothesis.
We generate eight hypotheses for each MRT step ($k $=8) with beam search.
We train with smoothed BLEU \citep{lin-och-2004-orange} from the Moses implementation.\footnote{\url{https://github.com/jwieting/beyond-bleu/blob/master/multi-bleu.perl}} Moreover, we use this metric to report results and verify they match sacrebleu \citep{post-2018-call}.\footnote{\url{https://github.com/mjpost/sacrebleu}}
%and report sacrebleu \citep{post-2018-call}\footnote{\url{https://github.com/mjpost/sacrebleu}}.
%\oa{it's better to use an equation environment than adding the equations inline. A: only does two?}
\paragraph{Optimization.}
We train the MLE objective over 200 epochs and
the combined RL objective over 15.
We perform early stopping by selecting the model with the
lowest validation loss.
We optimize with Nesterov's accelerated gradient method
\citep{sutskever2013importance} with a learning
rate of 0.25, a momentum of 0.99, and
re-normalize gradients to a 0.1 norm \citep{pascanu2012difficulty}.

\paragraph{Data.}
We experiment with four languages: German (De), Czech
(Cs), Russian (Ru), and Turkish (Tr), translating each of them to English (En). For training data for cs-en, de-en, and ru-en,
we use the WMT News Commentary v13\footnote{\url{http://data.statmt.org/wmt18/translation-task}}
\citep{bojar-etal-2017-findings}.
For tr-en training data, we use
WMT 2018 parallel data, which consists of the
SETIMES2 corpus \citep{Tiedemann2012ParallelDT}.
The validation set is a concatenation of
newsdev 2016 and 2017 released for WMT18. Test sets are the official WMT18 test sets.
Those experiments focus on a low-resource setting. We choose this setting as RL experiments are computationally demanding and this setting is common in the literature for RL experiments like ours \citet{wieting2019beyond}.
% Due to the computational cost of RL training, we train our models in a low-resource setting and add medium-size experiment in DE-EN only (see Supp. \S\ref{sec:app_entropy_stv_ltv}). 
%Those experiments focus on a low-resource setting. We choose this setting as RL experiments are computationally very demanding. We repeat this and further experiments with medium size data only for German and got similar results (see Supp. \S\ref{sec:app_entropy_stv_ltv}).
(see data statistics in Supp. \S\ref{sec:data_sizes})

% \paragraph{Optimization.}
% We optimize with Nesterov's accelerated gradient method
% \citep{sutskever2013importance} with a learning
% rate of 0.25, a momentum of 0.99, and
% re-normalize gradients to a 0.1 norm \citep{pascanu2012difficulty}.
% We train the MLE objective over 200 epochs and
% the combined RL objective over 15.
% We select the model with the
% lowest validation loss.

% \paragraph{Data.}

% We experiment with four languages: German (de), Czech
% (cs), Russian (ru), and Turkish (tr), translating each of them to English (en).
% For training data for cs-en, de-en, and ru-en,
% we use the WMT News Commentary v13\footnote{\url{ http://data.statmt.org/wmt18/translation-task}}
% \citep{ondrej2017findings} for training the models.
% Test sets are the official WMT18 test sets.
% For tr-en, we used for training the
% WMT 2018 parallel data, which consists of the
% SETIMES2 corpus \citep{Tiedemann2012ParallelDT}.
% The validation set is a concatenation of
% newsdev 2016 and 2017 released for WMT18.(see data statistics in Supp. \S\ref{sec:data_sizes})
%Test sets were the official WMT 2018 test sets.

\section{Reducing the Vocabulary Size} \label{sec:first} 

%RL for NLP suffers from long convergence time and other problems \citep[discussed in \S\ref{sec:back_rl_nmt},][]{choshen2019weaknesses} while it is quite successful in other domains \cite{Schulman2017ProximalPO}.
%The main difference between NMT and other tasks in which RL methods excel is the size of the action space, which in NMT includes all tokens in the vocabulary, typically tens of thousands of actions or more. Where there are many actions, and only a few are highly-rewarding, these actions are seldom sampled and hence seldom reinforced. The only exceptions to this rule are where the target action is initially favored by the policy (in which case, RL only re-ranks the top few actions, which dramatically limits its utility), or where the network can generalize across related actions. Hence we hypnotized that by decreasing the size of the space of allowed actions, we will enable better generalization across actions.

We begin by directly testing our hypothesis that the size of the action space is a cause for the long convergence time of RL for NMT. To do so, we train a model with target-side BPE taken from a much smaller vocabulary than is typically used.

We begin by training two MLE models, one with a large (17K-31K) target vocabulary (LTV) and another with a target vocabulary of size 1K (STV). The source vocabulary remains unchanged. We start with the MLE pretraining and then train each of the two models with RL.

Results (Table \ref{tab:table-small-big-results}) show that the RL training with STV achieves about $1$ BLEU point more than the RL training with LTV.\footnote{Preliminary experiments showed that altering the random seed changes the BLEU score by $\pm 0.01$ points.} For a comparison of the models' entropy see Supp. \S\ref{sec:app_entropy_stv_ltv}.
In order to verify that the improvement does not stem from the choice of $\alpha$ mixing RL and MLE (see Eq. \ref{fine-tuning objective}), we repeat the training for De-En with $\alpha \in \{0,1\}$, we find that $\alpha=0.3$ is superior to both. 
% alpha values of 0,1
%$training with only the MLE objective ($\alpha = 0$) and the RL objective ($\alpha = 1$). 
Moreover, RL improves STV more than LTV when training with only the RL objective ($\alpha = 1$). This indicates that RL training 
% is at least a partial cause for 
contributes to
the observed improvement.

% table 1:
\begin{table}[tb!]
\centering

\begin{tabular}{ @{}lrrrr@{}  }
\toprule
Model  & \multicolumn{1}{l}{DE-EN} & \multicolumn{1}{l}{CS-EN} & \multicolumn{1}{l}{RU-EN} & \multicolumn{1}{l}{TR-EN} \\ \midrule
LTV    & 25.07                     & 15.16                     & 16.67                     & 12.76                     \\
LTV+RL & 25.67                     & 15.33                     & 16.9                      & 12.98                     \\
Diff.  & 0.6                       & 0.17                      & 0.23                      & 0.22                      \\ \midrule
STV    & 21.83                     & 13.79                     & 14.63                     & 10.37                     \\
STV+RL & 23.23                     & 14.62                     & 15.73                     & 11.96                     \\
Diff.  & 1.4                       & 0.83                      & 1.1                       & 1.59                      \\ \bottomrule
\end{tabular}
\caption{\label{tab:table-small-big-results} BLEU scores for translating four languages to English using MLE pretraining followed by RL, and comparing a model with a large vocabulary (LTV) to a small one (STV). The top (bottom) block presents results for LTV (STV) with and without RL, and the difference between them (Diff.).
RL with STV gains more than 1 point more (on average) over the pre-trained MLE model, than RL with LTV.}
\end{table}

We next turn to analyze what tokens are responsible for the observed performance gain. Specifically, we examine whether reducing the vocabulary size resulted in RL being able to promote target tokens that received a low rank by the pre-trained MLE model. %(Failure to do so was one of the main criticisms of C19 against current RL practices in NMT)
%First, we denote by $y_{best}$\oa{didn't you use $y_i$ before for the target?} the token in $V_{T}$ to be the target token in a given context (source sentence $x$\oa{you used capital X before}, and prefix $y'_{<i}$) also known as a golden token. 
For each model, for 700K trials, we compute what rank the model assigns to the gold token $y_{i}$ for a context $y'_{<i}$ and source sentence $X$. Formally, $\forall r \in |V_T|$, $P^{r}_{model} = \frac{\#\{gold \ token \ assigns \ to \ the \ r \ rank\}}{\# \{all \ trials\}}$. %\oa{given gold std prefix, right? say so}, and plot a histogram of the rank of $y_{best}$.  
%We do so once for the pretrained model and once for RL model. 
%From this data, we build an empirical distribution over the $V_{T}$ possibilities, describing the model probability to assigns $y_{best}$ some position. 
%We can harness this distribution as a quantitative tool to assess model quality. For instance, we can expect a good model to concentrate most probability mass in a few of the first places as the model goal is to assign $y_{best}$ as first place.
We then compare the rank distribution of the MLE model to that of the RL model by subtracting those two distributions. In our notation, for each rank r, $\Delta P^r = P^{r}_{RL} - P^{r}_{MLE}$.
This subtraction represents how RL influences the model's ability to assign the correct token $y_{i}$ for each rank. The greater the positive effect of RL is, the more probable it is that the probability will be positive for the first rank, and negative for lower ranks (due to the probability shift to first place).

\begin{figure}[tb!]
\centering
\includegraphics[scale=0.33]{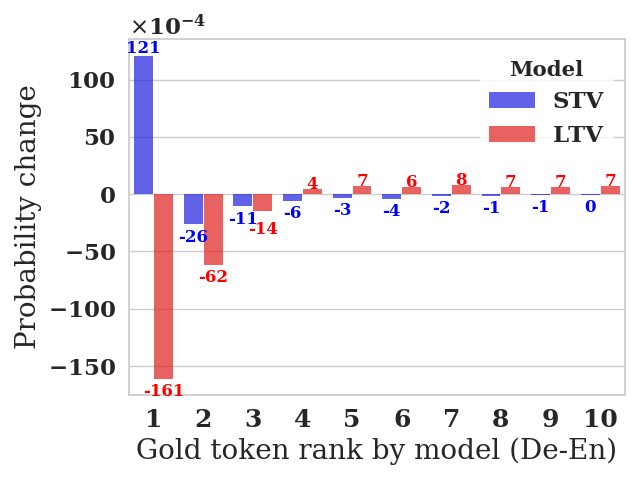}
\caption{Comparison of probability shift due to RL training of assign $y_{best}$ for ten first words for both LTV and STV. in blue, you can see the results with BPE of size 1,000, STV. in red are the results with BPE of size 30,000, LTV. a clear improvement of assigning $y_{best}$ in first place for STV.}
\label{fig:pro-shift-10-big-small}
\end{figure}

Figure \ref{fig:pro-shift-10-big-small} presents the probability difference per rank for LTV and STV. 
We can see that for the first rank the probability shift due to RL training with STV is more than twice the shift caused by RL training with LTV. Consequently, the probability shift for the following ranks is usually more negative for small vocabulary settings.
The figure indicates that indeed the shift of probability mass to higher positions occurs substantially more when we apply RL using a smaller action space. Moreover, the STV training was able to shift probability mass from lower ranks upwards compared to LTV. An indication for that is that, within the first one hundred ranks, STV reduces the probability of $83$ of them, whereas LTV of only $2$.
%See results for smaller model in Supp. \S\ref{sec:app_stv_ltv_small}.

% figure 1

%%%%%%%%%%%%%%%%%%%%%%%%%%%%%%%%%%%%%%%%%%%%%%%%%%%%%%%%
\section{Reducing the Effective Dimensionality of the Action Space}\label{sec:second}

%\oa{this section is good, but it's repetitious. please try to reorganize it}

Finding that reducing the number of actions improves RL's performance, we propose a method for reducing the effective number of actions, without changing the actual output.
%Reducing the vocabulary size might be undesirable due to out of domain words \citep{Koehn2017SixCF} and words distribution \citep{Gowda2020FindingTO} where we want domain specific words to be contain in the vocabulary, or where it is too costly to train a new model with small vocabulary.\oa{this last sent is unlcear; explain what you mean}
The vocabulary size might be static, as in pre-trained models \citep{Devlin_2019}, and reducing it might help RL but be sup-optimal for MLE \citep{Gowda2020FindingTO}, or introduce out-of-domain words \citep{Koehn2017SixCF}.
We propose to do so by using target embeddings that generalize over tokens that appear in similar contexts.
We explore two implementations of this idea, one where we initialize the target embeddings with high-quality embeddings, and another where we freeze the learned target embeddings during RL. We also explore a combination of the two approaches.
Freezing the target embeddings (%the
decoder's last layer) can be construed as training the network to output the activations of the penultimate layer, where a fixed function then maps it to the dimension of the vocabulary. 

We note that although freezing is a common procedure \citep{zoph2016transfer, Thompson_2018, lee2019elsa, DeCoster2021FrozenPT}, as far as we know, it has never been applied in the use of RL for sequence to sequence models. 
%In comparison with other methods for incorporating BERT into NMT \citep{Zhu2020Incorporating, yang2019towards, clinchant2019use}, our method is appealing as it does not require any architectural changes and is less computationally demanding.

%We may therefore think of the network as predicting an action in a continuous action space in the dimension of the inner representation of the network.
%We hypothesize that if $h$ assigns similar probabilities to actions (tokens) that appear in similar contexts, this may lead to performance gains without having to change the vocabulary size.

%We hypothesize that similar actions share parameters mean that the common parameters act as the actions rather than the end actions. Another way to enforce the use of shared parameters is to freeze the last layer. Thus, the output of the network is the last hidden layer and a fixed function then translates this layer to actions.

% This approach does not restrict the target vocabulary size, but the policy parameterization. 
Denote the function that the network computes with $f_{\theta}$.
$f_{\theta}$ can be written as $h_{\theta_{2}} \circ g_{\theta_{1}}$,
where $\theta = \left( \theta_{1}, \theta_{2} \right)$, $g$ maps the input -- source sentence $X$
and model translation prefix $y'_{<i}$ -- into $\mathbb{R}^d$, and $h$ maps $g$’s output into $\mathbb{R}^{|V_{T}|}$.

Using this notation, we can formulate the method as loading pre-trained MLE target embeddings to $h_{\theta_{2}}$ or freezing it (or both). 
As for many encoder-decoder architectures (including the Transformer), it holds that $d \ll |V_{T}|$, this can be thought of as constraining the agent to select a $d$-dimensional continuous action, where $h_{\theta_{2}}$ is a known transformation performed by the environment.

%The intuition behind the importance of the target embeddings is that it allows better generalization over related actions. Consider two actions which have similar meaning (e.g. synonyms). If $h$ respects this similarity, updates following experience with one action also provide a learning signal for the other action, which arises from changes to $g$. However, if $h$ is allowed to change as well, there could be an embedding ``drift'', since updates can change $h$ to directly affect the probability of the particular experienced action. Freezing $h$ would oppose even this embedding drift. The same intuition holds also for actions which are related but are very non-similar (e.g. antonyms), and for non-RL training. In the hypothetical case of a perfect $h$, the smaller layer (of dimension $d$) before the target embeddings acts as an effective continuous action, and is the semantic vector the network learns to predict.
The importance of target embedding is that they allow for better generalization over actions. The intuition is as follows.
%The intuition behind the importance of target embeddings is as follows.
Assume two tokens have the same embedding, and similar semantics, i.e., they are applicable in the same contexts (synonyms). Since they have the same target embeddings, during training the network will perform the same gradient updates when encountering either of them, except for in the last layer (since they are still considered different outputs).
%The network activations leading to their respective probabilities are identical, except for the (similar) target embeddings.
%Therefore, gradients towards one would lead to the other in all but the last similar but unshared layer. %Therefore, their similar initial embeddings would help generalizing. The last layer's update would change only the seen word. 
If the target embeddings are not frozen, encountering either of them during training will lead to very similar updates (since they have the same target embeddings), but their target embeddings may drift slightly apart, which will cause a subsequent drift in the lower layers. If the target embeddings are frozen, the gradient updates they will yield will remain the same and expedite learning. We hypothesize a similar effect during training, where tokens that have similar (but not identical) embeddings, and a similar (but not identical) distribution would benefit in training from each other. This motivates us to explore a combination of informative initialization and parameter freezing. (see formal proof in Supp. \S\ref{sec:app_proof}).

%In this (ideal) initialization case, the small layer before the target embeddings acts as the effective action space and is the semantic vector the network learns to predict. 

%By fixing the ultimate fully connected (FC) layer, RL is forced to determine the features that constitute the underlying lower-dimensional representation of the state, rather than modify the target embedding to conform to the representation learned by $g_{\theta_{1}}$. De facto, freezing the last FC layer forces the embedding space composed by this layer to be a static target embedding that the decoder is basing on through all RL training. 
%Therefore one can think of the model output as a semantic vector in the predefined frozen embedding space, and the RL adjusts that vector to maximize the reward function.

\subsection{Motivating Simulation through Policy Parameterization in Large Action Spaces}\label{sec:simulation}

In order to examine the intuition outlined above in a controlled setting, we consider a synthetic RL problem in which the action space is superficially enlarged.
The task is a (contextual) multi-armed-bandit, with $K$ actions. At each step, an input state is sampled from the environment (the "context"; a random vector sampled from a multivariate Gaussian distribution). A random, fixed, non-linear binary classifier determines whether action \#1 or action \#2 is rewarding based on the given context (actions $3$-$K$ are never rewarding), and the reward for each action is $r+z$ where $r=1$ for the rewarding action and $0$ for all other actions, and $z\sim\mathcal{N}\left(0,0.1\right)$. Crucially, we duplicate each action $a$ times, resulting in a total of $K\times a$ actions at the policy level -- whereas for the environment all `copies' of a given action are equivalent.

The problem structure, including the classifier itself, is unknown to the RL agent, which directly optimizes a policy parameterized as a fully-connected feed-forward neural network. We control two aspects of the last layer of the policy network, resulting in a total of four variants of agents. First, the last layer can be frozen to its initial value, or learned (by RL). Second, the last layer can be initialized at random, or induce a prior regarding the duplicated actions (such that weight vectors projecting to different copies of a given action are initialized identically). We call the latter the {\it informative} initialization.

\begin{figure}[tb!]
\centering
\includegraphics[width=\columnwidth,keepaspectratio]{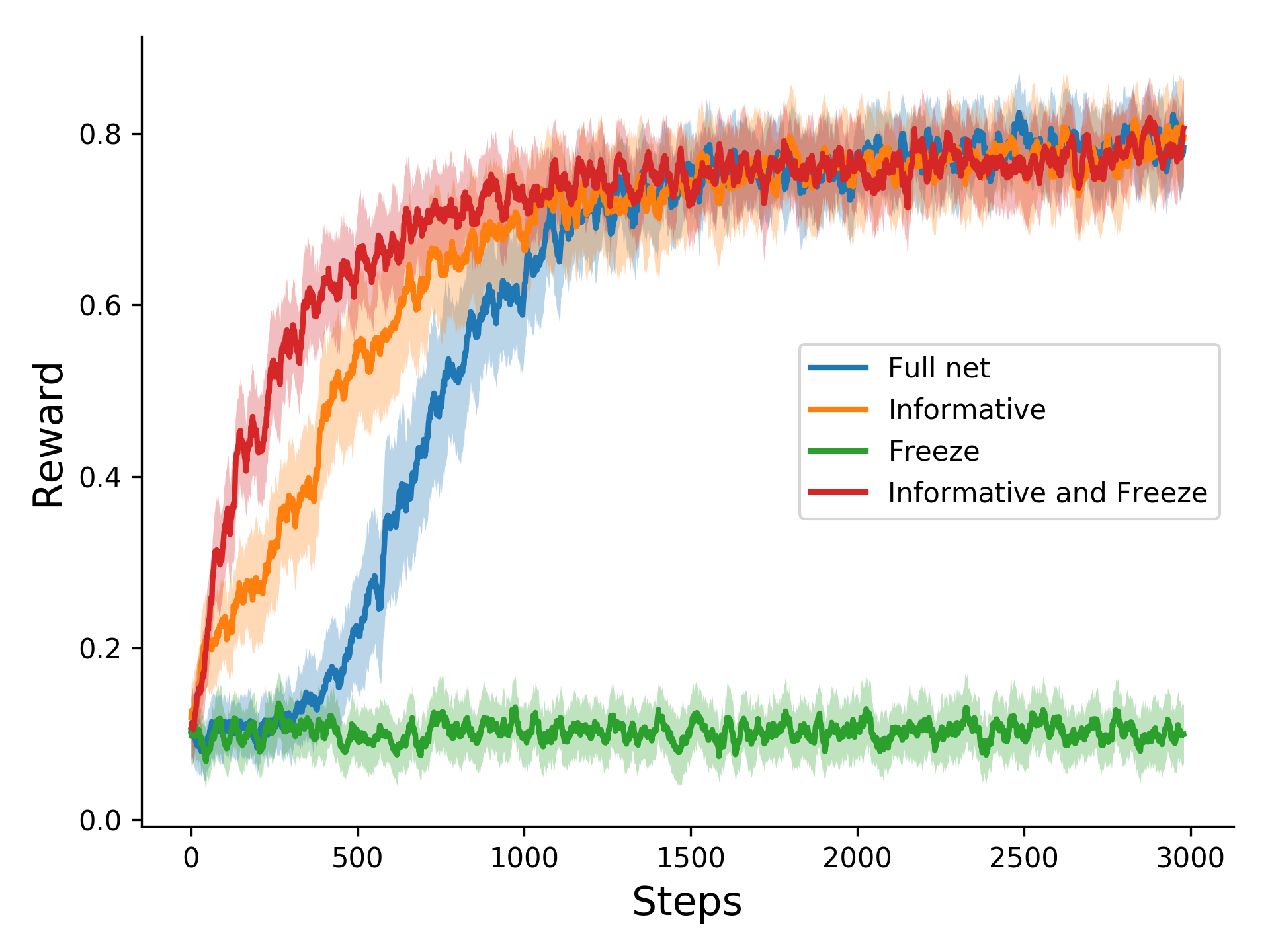}
\caption{Simulating learning in large action spaces. Figures show a moving average over $20$ steps of the underlying binary reward. Solid curves denote mean, shaded area denote $\pm0.5$ s.d. ($N=50$ trials per agent, $K=10,  a=400$, network architecture: 10-300-300-4000). Informative initialization is effective on its own, and more so when freezing is applied.\label{fig:duplicated_actions_reward}}
\end{figure}

We stress that the informative initialization carries no information about the underlying reward structure of the problem (i.e., the classifier, and the identity of the rewarding actions), but only as to which actions are duplicated. Nevertheless, as shown in Figure~\ref{fig:duplicated_actions_reward}, a prior regarding the structure of the action space is helpful on its own, leading to faster learning (compare \textit{Informative} to \textit{Full net}).

%Results fit the intuition presented. With informative last layer initialization, learning in previous layers naturally generalizes over the duplicated actions: if a rewarding action is sampled, changes in the activation patterns in the penultimate layer (caused by changes in earlier weights) increase the probabilities of all ``copies'' of this action together. Even if the last layer is allowed to change, it takes time for the weights to forget the initial (duplicated) condition. This implicit generalization over actions boosts early stages of learning, leading to faster convergence.
Results fit the intuition presented. With informative last layer initialization, learning in previous layers generalizes over the duplicated actions and boosts early stages of learning, leading to faster convergence.
We note that in this setting faster learning is not only the result of learning fewer parameters. Notably, freezing the last layer with random initialization, prohibits the network from learning the task. This is due to the regime of a very large action space (output layer; width $4000$) compared to the dimensionality of the hidden representations (width $300$). Freezing an informative initialization, on the other hand, sets the network in a rather different regime, in which the effective size of the output layer is (much) smaller than the hidden representation (i.e \#`real' actions; 10). In this regime, the network is generally expressive enough so that it can quickly learn the task even with a fixed, random readout layer \citep{soudry2018fix}.

To conclude, this example provides evidence that initializing and possibly freezing the last layer in the policy network in a way that respects the structure of the action space is helpful for learning in vast action spaces, as it supports generalization over similar or related actions. Importantly, this helps even when the (frozen) initialization does not contain task-specific information. In a more realistic scenario, actions are not simply a complete duplicate of each other, but rather are organized in some complex structure. Informative initialization, then, accounts not for duplicating weights, but for initializing them in such a way that a-priori reflects, or is congruent, with this structure. This motivates our approach -- in the realistic, complicated task of MT -- to freeze a learned output layer for the policy network, from a model whose embeddings have been shown to be effective across a wide range of tasks (in our case, BERT).

%\oa{can we change the legend in the figure to `informative' instead of `duplicate'?}

% These two novel approaches force adjustment to the MLE procedure. To the best of our knowledge, those adjustments too, has not been researched yet.
%These two new approaches are an unprecedented \oa{novel} path of research. Moreover, they force adjustment to the MLE procedure.  To the best of our knowledge, those adjustment has not been researched yet.
% Those alterations introduce new exacting results, as presented in \S\ref{Additional Experiments}.
% In the Experiment section (section \S\ref{sec:Experiments}), we will focus on the advancement of those two approaches to the performance of the RL mechanism.

%%%%%%%%%%%%%%%%%%%%%%%%%%%%%%%%%%%%%%%%%%%%%%%%%%%%%
\subsection{NMT Experiments} \label{sec:NMT_EXP}

%Although the first approach shows compelling results, it may be undesirable to restrict the target vocabulary in such a way due to a variety of practical reasons. For example, when translating medical documents, one may wish to represent unique terminology.  

The motivating analysis and simulations indicate that it is desirable to use target embeddings that assign similar values to similar actions. Doing so can be viewed as an effective reduction in the dimensionality of the action space. We turn to experiment with this approach on NMT. We explore two approaches: 
(1) freezing $h$ during RL; (2) informatively initializing $h$, as well as their combination. Our main results are presented in Table \ref{tab:table-BERT-results}.

As a baseline, we experiment with freezing uninformative target embeddings: target embeddings are randomly initialized and frozen during both MLE and RL. Unsurprisingly, doing so does not help training, and in fact, greatly degrades it (in about 2 BLEU points in En-De).

Next, we examine whether the target embeddings of the MLE pretraining are informative enough, namely whether freezing them during RL leads to improved effectiveness. Results show a slight improvement in BLEU when doing so, which is encouraging given that the frozen embeddings' weights consist of more than half of the network's trainable parameters. 
%Indeed, in the supplementary material we present the number of trainable parameters in each setting, showing that their number is cut by 61--63\% when freezing, yielding a more time-efficient RL step. 
Indeed, freezing the embedding layers has a dramatic impact on the volume of trainable parameters, decreasing their size by more than $60\%$. In Supp. \S\ref{sec:app_pram} we present the number of trainable parameters in each setting.
%Initial results (not shown) on German show that training MLE with frozen random embeddings hurts training badly, as in the simulation. MLE pre-training MT is usually used to direct the RL's exploration. We check if those target embeddings act similarly to the "duplicate" in the simulation.

We therefore hypothesize that, as in the simulations (\S\ref{sec:simulation}), the quality of the frozen embedding space is critical for the success of this approach. As using frozen MLE embeddings improves performance, but only somewhat, we further consider target embeddings that were trained on much larger datasets, specifically BERT's embedding 
layer.\footnote{\href{https://huggingface.co/}{HuggingFace} implementation}
%Therefore, we estimate that using a superior pre-trained embedding space would aid the RL more. We conduct a preliminary experiment aimed at examining the quality of BERT target embeddings.

% % table 2:
% \begin{table}[hb]
% \centering
% \begin{tabular}{ |p{2cm}|p{1cm}|p{1cm}|p{1cm}|p{1cm}|  }
%  \hline
%  Model & de-en & cs-en &ru-en& tr-en\\
%  \hline
%  MLE   & 26.18    & 19.38 & 18.09 & 14.34\\
%  RL&   30.18  & 19.42   & 18.46 & 17.12\\
%  RL+freezing & 30.13 & 19.5 &  18.57 & 17.21\\

%  \hline
% % \end{tabular}
% \caption{\label{tab:table-reg-freeze} BLEU results on translating four languages to English for MLE, RL and RL with freezing. The freezing results generally slightly outperform improvement over the regular RL training.} 
% \end{table}

%Observing that the BERT embeddings better share parameters across synonyms, we investigate the influence of initializing with BERT embeddings as target embeddings.
For this set of experiments, we adjust the target vocabulary to be BERT's vocabulary of size $|V_{T}| = 30526$. We train RL models with and without freezing the embedding layers and with and without loading BERT embedding. 
We report results of MLE training with BERT's embedding when the embedding is kept frozen as it reaches superior results
%\oa{what results are you talking about here? all the results you report? some of them? be explicit about this}
(see Supp. \S\ref{sec:app_bert_mle}).
%Notably, using BERT's vocabulary lowers the MLE results compared to the joint BPE vocabulary (\S\ref{subsec:Arch.}) which is known to be helpful, especially when source and target languages are close \citep{sennrich-etal-2016-neural}, this is aligned with our results (Table \ref{tab:table-BERT-results}). These considerations are peripheral to our discussion, which specifically targets the effectiveness of the RL approach.

% We train RL models with and without freezing the embedding layers and with and without loading BERT embedding. 
% We report results of MLE training with BERT's embedding when the embedding is kept frozen as it reaches superior results
% %\oa{what results are you talking about here? all the results you report? some of them? be explicit about this}
% (see Supp. \S\ref{sec:app_bert_mle}).

The results (Table \ref{tab:table-BERT-results}) directly parallels our findings in the simulations: Initializing from BERT ($+BERT+RL$) improves performance across all language pairs, and freezing ($+RL+FREEZE$) yields an additional improvement in most settings, albeit a more modest one. Combining both methods provides additional improvement. Indicating that FREEZE and BERT are helpful both independently and in conjunction. In total, our model ($+BERT+RL+FREEZE$) achieves $1.5$ BLEU points improvements over regular $RL$.
%Moreover, initializing with BERT is beneficial for MLE as well.
%Notably, while our focus is on RL methods, BERT also improves the MLE training.
% We also report semantic similarity scores in \ref{tab:my_label_sim}.
We also report semantic similarity scores in Supp. \S\ref{sec:app_bert_sim}.

Notably, our method surpasses the LTV+RL results (Table \ref{tab:table-small-big-results}) across all languages except German, overall about 1.5 BLEU points more on average. We hypothesize that the reason for this degradation is the lower results of the MLE with BERT's vocabulary compared to the joint BPE vocabulary which is known to be superior to BPE on each language individually, especially when the source and target languages are close \citep{sennrich-etal-2016-neural}. These considerations are peripheral to our discussion, which specifically targets the effectiveness of the RL approach.
% We report semantic similarity scores in Supp. \S\ref{sec:app_bert_sim}.

\begin{figure}[t!]
\centering
\includegraphics[scale=0.35]{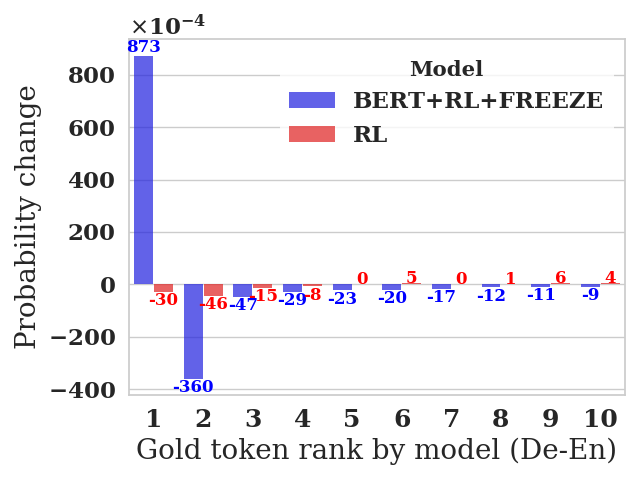}
\caption{Comparison of the change in the rank distribution of the target token following RL in two settings, one where RL training with frozen BERT embeddings is used (blue) and the second when we used basic RL training (red). The gain in probability in the first rank indicates that the model is more probable to be correct (which is reflected in its superior performance over the pre-trained MLE model). The negative values in the following places demonstrate how RL with frozen high-quality target embeddings can improve not only when the MLE model is initially close to being correct.}
\label{fig:freezing embedding improvment - BERT}
\end{figure}

%Moreover, the models with frozen embedding outperform the models without freezing in terms of BLEU score in most cases (see table \ref{tab:table-BERT-results}). 
Finally, initializing from BERT increases RL's ability to promote tokens that were not ranked high according to the MLE model (Fig.~\ref{fig:freezing embedding improvment - BERT}).
%Amongst the first 100 ranks, STV reduces the probability of $97$ of them, whereas LTV  of only $5$.

%\oa{is it de-en? also in fig. 1? in both indicate the language pair you're working on}
%As before, freezing the embedding layers has a dramatic impact on the volume of trainable parameters, decreasing their size by more than $60\%$ (see Supp. \S\ref{sec:app_pram}).
%To conclude, initializing from BERT's embeddings and freezing them seems to improve both RL's effectiveness and efficiency.

% % table 5:
% \begin{table}[ht]
% \centering
% \begin{tabular}{p{3cm}p{1cm}p{1cm}p{1cm}p{1cm}  }
%  \toprule
%   Model & de-en & cs-en &ru-en& tr-en\\
%  \midrule
%  MLE with BERT  & 23.46    & 16.59 & 18.14 & 14.15\\
%  RL without freeze &   24.44  & 17.04   & 18.68 & 14.37\\
%  RL with freeze & 24.71 & 17.37 &  18.3 & 14.55\\
%  \bottomrule
% \end{tabular}

% \end{table}

\begin{table}
\large
\resizebox{\columnwidth}{!}{%
\begin{tabular}{@{}lllll@{}}
\toprule
MODEL           & De-En & Cs-En & Ru-En & Rr-En \\ \midrule
MLE             & 22.38 & 15.81 & 17.31 & 12.60 \\
+RL             & 23.19 & 15.81 & 17.31 & 12.66 \\
% \hline
\midrule

% \hline
% \hline
+RL+FREEZE      & 23.14 & 16.04 & 17.78 & 13.18 \\
% \midrule
+BERT           & 23.46 & 16.59 & 18.14 & 14.15 \\
+BERT+RL       & 24.44 & 17.04 & \textbf{18.68} & 14.37 \\
+BERT+RL+FREEZE & \textbf{24.71} & \textbf{17.37} & 18.30 & \textbf{14.55} \\ \bottomrule
\end{tabular}%
}\caption{\label{tab:table-BERT-results}BLEU scores on translating four languages to English.
The upper block shows the baseline scores of training only with MLE, and with MLE followed by RL. RL presents modest improvement (if any) over only using MLE.
+RL+FREEZE shows some improvement due to freezing the target embeddings.
The lower three rows show results when using BERT's target embeddings (informative initialization).
%RL is more effective in this setting (the difference between +RL+BERT and +BERT is greater than the difference between +RL and MLE). 
Additional benefit is seen from freezing (+RL+BERT+FREEZE). }
% with three modification to the training over the MLE baseline, RL - RL training phase, FREEZE - freezing the embedding layer and BERT - loading bert embedding. 
% we show all the intermediate results of training modification from RL, FREEZE and BERT. in three out of four languages the best results is from the combination of the three modifications.}

%\label{tab:table-reg-freeze}
%BLEU results on translating four languages to English for MLE, RL and RL with freezing. The freezing results generally slightly outperform improvement over the regular RL training.}
\end{table}

\section{Human Evaluation}\label{sec:human_eval}

\begin{figure}[t!]
\centering
\includegraphics[scale=0.34]{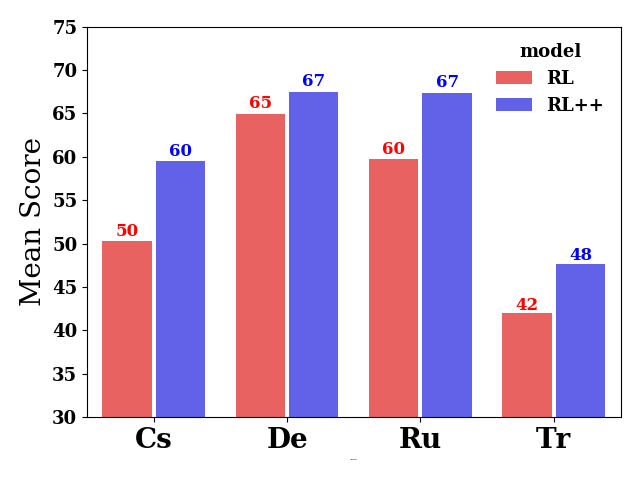}
\caption{Average human ratings on 200 sentences from
the test set for each of the respective languages. $RL$ is the baseline RL model and $RL++$ is our model ($+BERT+RL+FREEZE$). The performance of our model is consistently better than the baseline.\vspace{-.2cm}}
\label{fig:human_eval}
\end{figure}

We perform human evaluation, comparing
the baseline $RL$ with our proposed model. We selected 100
translations from the respective test sets of each language. The annotation was performed by two professional annotators (contractors of the project), who work in the field of translation. Both are native English speakers. The annotators assigned a
score from 0 to 100, judging how well the translation conveyed the information contained in the reference (see annotation guidelines in Supp. \S\ref{sec:human_eval_info}).
From Fig. \ref{fig:human_eval}, we see that our proposed model scores the highest
across all language pairs.
To test statistical significance, we use the Wilcoxon rank sum test to standardize score distributions fit for our setting \citep{Graham2015CanMT}. Comparing the two models' distributions, we got a p-value of $8.5e^{-5}$
%\oa{on which comparison did you get this p value? comparing what and what?}
indicating the improvement is significant.
We emphasize that our main goal is showing that our method can improve the optimized metric (e.g., BLEU), and hence the improvement over the semantic similarity score and the human evaluation is an additional indication of our method's robustness. 

% figure 1

\section{Comparing Target Embedding Spaces}\label{seq:comp_embd}

The previous section discussed how BERT's target embeddings improve RL performance, compared to target embeddings learned by MLE. We now turn to directly analyze the generalization ability of the two embeddings. We do so by comparing the embeddings of semantically related words.

\begin{figure}[t!]
\centering
\begin{subfigure}{0.5\textwidth}
\includegraphics[width=0.9\linewidth, height=5cm]{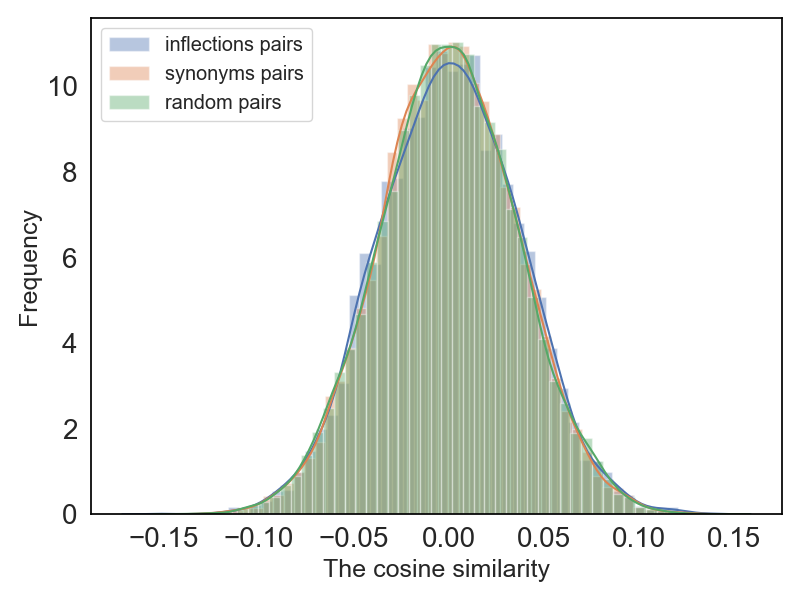} 
\caption{MLE embeddings}
\label{fig:sub.im.2.1}
\end{subfigure}
\begin{subfigure}{0.5\textwidth}
\includegraphics[width=0.9\linewidth, height=5cm]{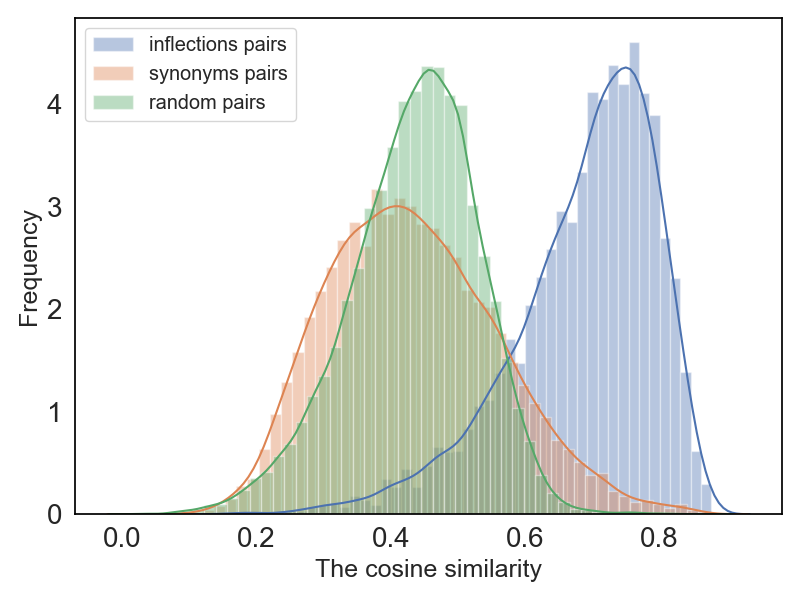}
\caption{BERT embeddings}
\label{fig:sub.im.2.2}
\end{subfigure}
\caption{Comparison of the distribution of the cosine similarity between word pairs from three groups: random word pairs (in green), synonym pairs that do not share a stem (in orange), and pairs of synonyms that share a stem  (in blue). The top figure refers to the target embeddings learned by MLE, and the bottom one to BERT embeddings. The ability of the embeddings to distinguish between these three groups is informative of their ability to map semantically related words to similar embeddings. The better discrimination ability of BERT embeddings is thus likely related to their superiority as target embeddings over MLE embeddings.}
\label{fig:embedding Comparison}
\end{figure}

We use WordNet~\citep{miller1998wordnet} and spaCy\footnote{\url{https://spacy.io/}} to compile three lists of word pairs: inflections (e.g., 'documentaries' / 'documentary', 'boxes' / 'box', 'stemming' / 'stem'), synonyms (e.g., 'luckily' / 'fortunately', 'amazement' / 'astonishment', 'purposely' / 'intentionally'), and random pairs, and compare the embeddings assigned to these pairs using BERT and MLE embeddings. Figure \ref{fig:embedding Comparison} presents the distributions of the cosine similarity of the pairs in the three lists for both embedding spaces. Results show that MLE embeddings for the different lists have almost identical distributions, demonstrating the limited informativeness of these target embeddings. In contrast, BERT embeddings only display a small overlap between the similarity distributions of inflections and random pairs. However, synonyms' distribution remains quite similar to that of random pairs. In conclusion, BERT embeddings better discern semantics overall compared to MLE embeddings, which may partly account for their superior performance. Results also indicate BERT's embeddings could be further improved.

% figure 2:

% figure 3:
% \begin{figure}[ht]
% \begin{subfigure}{0.5\textwidth}
% \includegraphics[width=0.9\linewidth, height=5cm]{cs_p_c.png} 
% \caption{probability change in MRT for cs-en}
% \label{fig:sub.im.3.1}
% \end{subfigure}
% \begin{subfigure}{0.5\textwidth}
% \includegraphics[width=0.9\linewidth, height=5cm]{de_p_c.png}
% \caption{probability change in MRT for de-en}
% \label{fig:sub.im.3.2}
% \end{subfigure}

% \begin{figure}[ht]
% \begin{subfigure}{0.5\textwidth}
% \includegraphics[width=0.9\linewidth, height=5cm]{bert+rl+freeze_vs.rl.png}
% \caption{probability change in MRT for de-en}
% \label{fig:sub.im.3.2}
% \end{subfigure}

% \begin{figure}[!ht]
% \centering
% % \includegraphics[scale=0.28]{bert+rl+freeze_vs.rl_with_lag.png}
% \includegraphics[scale=0.35]{bert_prob_wMB.png}
% \caption{Comparison of the change in the rank distribution of the target token following RL in two settings, one where RL training with frozen BERT embeddings is used (blue) and the second when we used basic RL training (red). The gain in probability in the first rank indicates that the model is more probable to be correct (which is reflected in its superior performance over the pre-trained MLE model). The negative values in the following places demonstrate how RL with frozen high-quality target embeddings can improve not only when the MLE model is initially close to being correct.}
% \label{fig:freezing embedding improvment - BERT}
% \end{figure}

\section{Conclusion}

In this paper, we addressed the limited effectiveness of RL for NMT, seeking to understand its origins and offer means for tackling it. We hypothesized that this limitation arises from the size of the action spaces used in NMT and examined two ways of reducing their effective dimension. 
In the first method, we experiment with smaller vocabularies, showing improved RL effectiveness. While this method constrains the size of the vocabulary, which may be limiting in some settings \citep{ding2019call,Gowda2020FindingTO}, it motivates further research along these lines.
%the strong results we obtain may justify its use in real world settings.

The second approach introduces a new method of using informative target embeddings and potentially freezing them during RL. 
We find that this method may be beneficial as well, but its effectiveness crucially depends on the quality of the employed embeddings. Indeed, we find using both simulations and NMT experiments that freezing in itself results in some improvement in RL performance, but that combined with target embeddings that generalize over words with a similar distribution, it may yield substantial gains as shown by BLEU, semantic similarity, and human evaluation. We compare the target embeddings produced by MLE and those by BERT, finding the latter to be considerably stronger.
Those results in low resources settings, encourage further research aiming to address the problem of large action space for TG in richer data settings by adapting and extending our methods.

Future work will increase the exploration ability of RL training in NMT. A promising line of research towards this goal is using off-policy methods. Off-policy methods, in which observations are sampled from a different policy than the one we currently optimize, are prominent in RL \citep{watkins1992q, sutton1998introduction}, and were also studied in the context of policy gradient methods \citep{degris2012off, silver2014dpg}.
%Such methods allow learning from a more "exploratory" policy and allow smoothing.
We believe that the adoption of such methods to enhance exploration, combined with our proposed method for using target embeddings, can be a promising path forward for the application of RL in NMT, and more generally in TG.

A different line of future work will focus on changing the network's architecture to predict a $d$ dimension continuous action, instead of discrete actions. Such an approach may directly reduce the size of the action space without limiting the number of words that can be predicted.
%\lc{Because it will actually learn the penultimate layer representation, directly.}
%a variant of the freezing method. In this variant, the network's output will have the same dimension as the network-wide. The predicted token will be then defined as the token most similar to the output vector according to some predefined embedding. This variant has the advantage of leveraging existing embeddings without the need to transfer them as the model embeddings. This variant can also boost model agility and enable deployment of RL techniques that work for a model that predicts a vector in the target embedding \citep{dulacarnold2016deep}.

\section{Acknowledgements}
This work was supported by the Israel Science Foundation (grant no. 2424/21), and by the Applied Research in Academia Program of the Israel Innovation Authority.
\bibliography{anthology,custom}

\begin{thebibliography}{49}
\expandafter\ifx\csname natexlab\endcsname\relax\def\natexlab#1{#1}\fi

\bibitem[{Bojar et~al.(2017)Bojar, Chatterjee, Federmann, Graham, Haddow,
  Huang, Huck, Koehn, Liu, Logacheva, Monz, Negri, Post, Rubino, Specia, and
  Turchi}]{bojar-etal-2017-findings}
Ond{\v{r}}ej Bojar, Rajen Chatterjee, Christian Federmann, Yvette Graham, Barry
  Haddow, Shujian Huang, Matthias Huck, Philipp Koehn, Qun Liu, Varvara
  Logacheva, Christof Monz, Matteo Negri, Matt Post, Raphael Rubino, Lucia
  Specia, and Marco Turchi. 2017.
\newblock \href {https://doi.org/10.18653/v1/W17-4717} {Findings of the 2017
  conference on machine translation ({WMT}17)}.
\newblock In \emph{Proceedings of the Second Conference on Machine
  Translation}, pages 169--214, Copenhagen, Denmark. Association for
  Computational Linguistics.

\bibitem[{Caccia et~al.(2018)Caccia, Caccia, Fedus, Larochelle, Pineau, and
  Charlin}]{caccia2018language}
Massimo Caccia, Lucas Caccia, William Fedus, Hugo Larochelle, Joelle Pineau,
  and Laurent Charlin. 2018.
\newblock Language gans falling short.
\newblock \emph{arXiv preprint arXiv:1811.02549}.

\bibitem[{Chandak et~al.(2019)Chandak, Theocharous, Kostas, Jordan, and
  Thomas}]{Chandak2019LearningAR}
Yash Chandak, Georgios Theocharous, James Kostas, Scott~M. Jordan, and P.~S.
  Thomas. 2019.
\newblock Learning action representations for reinforcement learning.
\newblock \emph{ArXiv}, abs/1902.00183.

\bibitem[{Choshen et~al.(2019)Choshen, Fox, Aizenbud, and
  Abend}]{choshen2019weaknesses}
Leshem Choshen, Lior Fox, Zohar Aizenbud, and Omri Abend. 2019.
\newblock On the weaknesses of reinforcement learning for neural machine
  translation.
\newblock In \emph{International Conference on Learning Representations}.

\bibitem[{Coster et~al.(2021)Coster, D’Oosterlinck, Pizurica, Rabaey,
  Verlinden, Herreweghe, and Dambre}]{DeCoster2021FrozenPT}
Mathieu~De Coster, Karel D’Oosterlinck, Marija Pizurica, Paloma Rabaey,
  Severine Verlinden, Mieke~Van Herreweghe, and Joni Dambre. 2021.
\newblock Frozen pretrained transformers for neural sign language translation.
\newblock In \emph{MTSUMMIT}.

\bibitem[{Degris et~al.(2012)Degris, White, and Sutton}]{degris2012off}
Thomas Degris, Martha White, and Richard~S Sutton. 2012.
\newblock Off-policy actor-critic.
\newblock \emph{arXiv preprint arXiv:1205.4839}.

\bibitem[{Devlin et~al.(2019)Devlin, Chang, Lee, and Toutanova}]{Devlin_2019}
Jacob Devlin, Ming-Wei Chang, Kenton Lee, and Kristina Toutanova. 2019.
\newblock \href {https://doi.org/10.18653/v1/n19-1423} {Bert: Pre-training of
  deep bidirectional transformers for language understanding}.
\newblock \emph{Proceedings of the 2019 Conference of the North}.

\bibitem[{Ding et~al.(2019)Ding, Renduchintala, and Duh}]{ding2019call}
Shuoyang Ding, Adithya Renduchintala, and Kevin Duh. 2019.
\newblock A call for prudent choice of subword merge operations.
\newblock \emph{CoRR, vol. abs/1905.10453}.

\bibitem[{Dulac-Arnold et~al.(2016)Dulac-Arnold, Evans, van Hasselt, Sunehag,
  Lillicrap, Hunt, Mann, Weber, Degris, and Coppin}]{dulacarnold2016deep}
Gabriel Dulac-Arnold, Richard Evans, Hado van Hasselt, Peter Sunehag, Timothy
  Lillicrap, Jonathan Hunt, Timothy Mann, Theophane Weber, Thomas Degris, and
  Ben Coppin. 2016.
\newblock \href {http://arxiv.org/abs/1512.07679} {Deep reinforcement learning
  in large discrete action spaces}.

\bibitem[{Edunov et~al.(2018)Edunov, Ott, Auli, Grangier, and
  Ranzato}]{Edunov_2018}
Sergey Edunov, Myle Ott, Michael Auli, David Grangier, and Marc’Aurelio
  Ranzato. 2018.
\newblock \href {https://doi.org/10.18653/v1/n18-1033} {Classical structured
  prediction losses for sequence to sequence learning}.
\newblock \emph{Proceedings of the 2018 Conference of the North American
  Chapter of the Association for Computational Linguistics: Human Language
  Technologies, Volume 1 (Long Papers)}.

\bibitem[{Gehring et~al.(2017)Gehring, Auli, Grangier, Yarats, and
  Dauphin}]{gehring2017convolutional}
Jonas Gehring, Michael Auli, David Grangier, Denis Yarats, and Yann~N. Dauphin.
  2017.
\newblock \href {http://arxiv.org/abs/1705.03122} {Convolutional sequence to
  sequence learning}.

\bibitem[{Gowda and May(2020)}]{Gowda2020FindingTO}
Thamme Gowda and Jonathan May. 2020.
\newblock Finding the optimal vocabulary size for neural machine translation.
\newblock In \emph{EMNLP}.

\bibitem[{Graham et~al.(2015)Graham, Baldwin, Moffat, and
  Zobel}]{Graham2015CanMT}
Yvette Graham, Timothy Baldwin, Alistair Moffat, and Justin Zobel. 2015.
\newblock Can machine translation systems be evaluated by the crowd alone.
\newblock \emph{Natural Language Engineering}, 23:3 -- 30.

\bibitem[{Hoffer et~al.(2018)Hoffer, Hubara, and Soudry}]{soudry2018fix}
Elad Hoffer, Itay Hubara, and Daniel Soudry. 2018.
\newblock \href {https://openreview.net/forum?id=S1Dh8Tg0-} {Fix your
  classifier: the marginal value of training the last weight layer}.
\newblock In \emph{International Conference on Learning Representations}.

\bibitem[{Koehn and Knowles(2017)}]{Koehn2017SixCF}
Philipp Koehn and R.~Knowles. 2017.
\newblock Six challenges for neural machine translation.
\newblock In \emph{NMT@ACL}.

\bibitem[{Lee et~al.(2019)Lee, Tang, and Lin}]{lee2019elsa}
Jaejun Lee, Raphael Tang, and Jimmy Lin. 2019.
\newblock \href {http://arxiv.org/abs/1911.03090} {What would elsa do? freezing
  layers during transformer fine-tuning}.

\bibitem[{Lin and Och(2004)}]{lin-och-2004-orange}
Chin-Yew Lin and Franz~Josef Och. 2004.
\newblock \href {https://www.aclweb.org/anthology/C04-1072} {{ORANGE}: a method
  for evaluating automatic evaluation metrics for machine translation}.
\newblock In \emph{{COLING} 2004: Proceedings of the 20th International
  Conference on Computational Linguistics}, pages 501--507, Geneva,
  Switzerland. COLING.

\bibitem[{Miller(1998)}]{miller1998wordnet}
George~A Miller. 1998.
\newblock \emph{WordNet: An electronic lexical database}.
\newblock MIT press.

\bibitem[{Mnih et~al.(2013)Mnih, Kavukcuoglu, Silver, Graves, Antonoglou,
  Wierstra, and Riedmiller}]{Mnih2013PlayingAW}
V.~Mnih, K.~Kavukcuoglu, D.~Silver, A.~Graves, Ioannis Antonoglou, Daan
  Wierstra, and Martin~A. Riedmiller. 2013.
\newblock Playing atari with deep reinforcement learning.
\newblock \emph{ArXiv}, abs/1312.5602.

\bibitem[{Och(2003)}]{och2003minimum}
Franz~Josef Och. 2003.
\newblock Minimum error rate training in statistical machine translation.
\newblock In \emph{Proceedings of the 41st annual meeting of the Association
  for Computational Linguistics}, pages 160--167.

\bibitem[{Ott et~al.(2019)Ott, Edunov, Baevski, Fan, Gross, Ng, Grangier, and
  Auli}]{ott2019fairseq}
Myle Ott, Sergey Edunov, Alexei Baevski, Angela Fan, Sam Gross, Nathan Ng,
  David Grangier, and Michael Auli. 2019.
\newblock fairseq: A fast, extensible toolkit for sequence modeling.
\newblock In \emph{Proceedings of the 2019 Conference of the North American
  Chapter of the Association for Computational Linguistics (Demonstrations)},
  pages 48--53.

\bibitem[{Papineni et~al.(2002)Papineni, Roukos, Ward, and
  Zhu}]{papineni2002bleu}
Kishore Papineni, Salim Roukos, Todd Ward, and Wei-Jing Zhu. 2002.
\newblock Bleu: a method for automatic evaluation of machine translation.
\newblock In \emph{Proceedings of the 40th annual meeting of the Association
  for Computational Linguistics}, pages 311--318.

\bibitem[{Pascanu et~al.(2012)Pascanu, Mikolov, and
  Bengio}]{pascanu2012difficulty}
Razvan Pascanu, Tomas Mikolov, and Yoshua Bengio. 2012.
\newblock \href {http://arxiv.org/abs/1211.5063} {On the difficulty of training
  recurrent neural networks}.

\bibitem[{Pereyra et~al.(2017)Pereyra, Tucker, Chorowski, Kaiser, and
  Hinton}]{Pereyra2017RegularizingNN}
G.~Pereyra, G.~Tucker, J.~Chorowski, L.~Kaiser, and Geoffrey~E. Hinton. 2017.
\newblock Regularizing neural networks by penalizing confident output
  distributions.
\newblock \emph{ArXiv}, abs/1701.06548.

\bibitem[{Post(2018)}]{post-2018-call}
Matt Post. 2018.
\newblock \href {https://doi.org/10.18653/v1/W18-6319} {A call for clarity in
  reporting {BLEU} scores}.
\newblock In \emph{Proceedings of the Third Conference on Machine Translation:
  Research Papers}, pages 186--191, Brussels, Belgium. Association for
  Computational Linguistics.

\bibitem[{Ranzato et~al.(2016)Ranzato, Chopra, Auli, and
  Zaremba}]{Ranzato2016SequenceLT}
Marc'Aurelio Ranzato, S.~Chopra, M.~Auli, and W.~Zaremba. 2016.
\newblock Sequence level training with recurrent neural networks.
\newblock \emph{CoRR}, abs/1511.06732.

\bibitem[{Ren et~al.(2019)Ren, Zhang, Liu, Zhou, and
  Ma}]{Ren2019UnsupervisedNM}
Shuo Ren, Zhirui Zhang, Shujie Liu, M.~Zhou, and S.~Ma. 2019.
\newblock Unsupervised neural machine translation with smt as posterior
  regularization.
\newblock \emph{ArXiv}, abs/1901.04112.

\bibitem[{Schulz et~al.(2018)Schulz, Aziz, and Cohn}]{schulz2018stochastic}
Philip Schulz, Wilker Aziz, and Trevor Cohn. 2018.
\newblock A stochastic decoder for neural machine translation.
\newblock \emph{arXiv preprint arXiv:1805.10844}.

\bibitem[{Sennrich et~al.(2016)Sennrich, Haddow, and
  Birch}]{sennrich-etal-2016-neural}
Rico Sennrich, Barry Haddow, and Alexandra Birch. 2016.
\newblock \href {https://doi.org/10.18653/v1/P16-1162} {Neural machine
  translation of rare words with subword units}.
\newblock In \emph{Proceedings of the 54th Annual Meeting of the Association
  for Computational Linguistics (Volume 1: Long Papers)}, pages 1715--1725,
  Berlin, Germany. Association for Computational Linguistics.

\bibitem[{Shah et~al.(2018)Shah, Hakkani-T{\"u}r, Liu, and
  T{\"u}r}]{Shah2018BootstrappingAN}
Pararth Shah, Dilek~Z. Hakkani-T{\"u}r, Bing Liu, and G.~T{\"u}r. 2018.
\newblock Bootstrapping a neural conversational agent with dialogue self-play,
  crowdsourcing and on-line reinforcement learning.
\newblock In \emph{NAACL-HLT}.

\bibitem[{Sharma et~al.(2017)Sharma, Suresh, Ramesh, and
  Ravindran}]{sharma2017learning}
Sahil Sharma, Aravind Suresh, Rahul Ramesh, and Balaraman Ravindran. 2017.
\newblock Learning to factor policies and action-value functions: Factored
  action space representations for deep reinforcement learning.
\newblock \emph{arXiv preprint arXiv:1705.07269}.

\bibitem[{Shen et~al.(2015)Shen, Cheng, He, He, Wu, Sun, and
  Liu}]{shen2015minimum}
Shiqi Shen, Yong Cheng, Zhongjun He, Wei He, Hua Wu, Maosong Sun, and Yang Liu.
  2015.
\newblock Minimum risk training for neural machine translation.
\newblock \emph{arXiv preprint arXiv:1512.02433}.

\bibitem[{Shen et~al.(2016)Shen, Cheng, He, He, Wu, Sun, and
  Liu}]{shen-etal-2016-minimum}
Shiqi Shen, Yong Cheng, Zhongjun He, Wei He, Hua Wu, Maosong Sun, and Yang Liu.
  2016.
\newblock \href {https://doi.org/10.18653/v1/P16-1159} {Minimum risk training
  for neural machine translation}.
\newblock In \emph{Proceedings of the 54th Annual Meeting of the Association
  for Computational Linguistics (Volume 1: Long Papers)}, pages 1683--1692,
  Berlin, Germany. Association for Computational Linguistics.

\bibitem[{Silver et~al.(2014)Silver, Lever, Heess, Degris, Wierstra, and
  Riedmiller}]{silver2014dpg}
David Silver, Guy Lever, Nicolas Heess, Thomas Degris, Daan Wierstra, and
  Martin Riedmiller. 2014.
\newblock \href {http://proceedings.mlr.press/v32/silver14.html} {Deterministic
  policy gradient algorithms}.
\newblock In \emph{Proceedings of the 31st International Conference on Machine
  Learning}, volume~32 of \emph{Proceedings of Machine Learning Research},
  pages 387--395, Bejing, China. PMLR.

\bibitem[{Sutskever et~al.(2013)Sutskever, Martens, Dahl, and
  Hinton}]{sutskever2013importance}
Ilya Sutskever, James Martens, George Dahl, and Geoffrey Hinton. 2013.
\newblock On the importance of initialization and momentum in deep learning.
\newblock In \emph{International conference on machine learning}, pages
  1139--1147.

\bibitem[{Sutton et~al.(1998)Sutton, Barto et~al.}]{sutton1998introduction}
Richard~S Sutton, Andrew~G Barto, et~al. 1998.
\newblock \emph{Introduction to reinforcement learning}, volume 135.
\newblock MIT press Cambridge.

\bibitem[{Szegedy et~al.(2016)Szegedy, Vanhoucke, Ioffe, Shlens, and
  Wojna}]{Szegedy_2016}
Christian Szegedy, Vincent Vanhoucke, Sergey Ioffe, Jon Shlens, and Zbigniew
  Wojna. 2016.
\newblock \href {https://doi.org/10.1109/cvpr.2016.308} {Rethinking the
  inception architecture for computer vision}.
\newblock \emph{2016 IEEE Conference on Computer Vision and Pattern Recognition
  (CVPR)}.

\bibitem[{Tennenholtz and Mannor(2019)}]{Tennenholtz2019TheNL}
Guy Tennenholtz and Shie Mannor. 2019.
\newblock The natural language of actions.
\newblock In \emph{ICML}.

\bibitem[{Thompson et~al.(2018)Thompson, Khayrallah, Anastasopoulos, McCarthy,
  Duh, Marvin, McNamee, Gwinnup, Anderson, and Koehn}]{Thompson_2018}
Brian Thompson, Huda Khayrallah, Antonios Anastasopoulos, Arya~D. McCarthy,
  Kevin Duh, Rebecca Marvin, Paul McNamee, Jeremy Gwinnup, Tim Anderson, and
  Philipp Koehn. 2018.
\newblock \href {https://doi.org/10.18653/v1/w18-6313} {Freezing subnetworks to
  analyze domain adaptation in neural machine translation}.
\newblock \emph{Proceedings of the Third Conference on Machine Translation:
  Research Papers}.

\bibitem[{Tiedemann(2012)}]{Tiedemann2012ParallelDT}
J.~Tiedemann. 2012.
\newblock Parallel data, tools and interfaces in opus.
\newblock In \emph{LREC}.

\bibitem[{Todorov et~al.(2012)Todorov, Erez, and Tassa}]{todorov2012mujoco}
Emanuel Todorov, Tom Erez, and Yuval Tassa. 2012.
\newblock Mujoco: A physics engine for model-based control.
\newblock In \emph{2012 IEEE/RSJ International Conference on Intelligent Robots
  and Systems}, pages 5026--5033. IEEE.

\bibitem[{Wang and Sennrich(2020)}]{wang2020exposure}
Chaojun Wang and Rico Sennrich. 2020.
\newblock \href {http://arxiv.org/abs/2005.03642} {On exposure bias,
  hallucination and domain shift in neural machine translation}.

\bibitem[{Watkins and Dayan(1992)}]{watkins1992q}
Christopher~JCH Watkins and Peter Dayan. 1992.
\newblock Q-learning.
\newblock \emph{Machine learning}, 8(3-4):279--292.

\bibitem[{Wieting et~al.(2019)Wieting, Berg-Kirkpatrick, Gimpel, and
  Neubig}]{wieting2019beyond}
John Wieting, Taylor Berg-Kirkpatrick, Kevin Gimpel, and Graham Neubig. 2019.
\newblock \href {https://arxiv.org/abs/1909.06694} {Beyond bleu: Training
  neural machine translation with semantic similarity}.
\newblock In \emph{Proceedings of the Association for Computational
  Linguistics}.

\bibitem[{Williams(1992)}]{Williams92simplestatistical}
Ronald~J. Williams. 1992.
\newblock Simple statistical gradient-following algorithms for connectionist
  reinforcement learning.
\newblock In \emph{Machine Learning}, pages 229--256.

\bibitem[{Wu et~al.(2018{\natexlab{a}})Wu, Tian, Qin, Lai, and Liu}]{Wu2018ASO}
Lijun Wu, Fei Tian, T.~Qin, J.~Lai, and T.~Liu. 2018{\natexlab{a}}.
\newblock A study of reinforcement learning for neural machine translation.
\newblock In \emph{EMNLP}.

\bibitem[{Wu et~al.(2018{\natexlab{b}})Wu, Tian, Qin, Lai, and
  Liu}]{wu2018study}
Lijun Wu, Fei Tian, Tao Qin, Jianhuang Lai, and Tie-Yan Liu.
  2018{\natexlab{b}}.
\newblock A study of reinforcement learning for neural machine translation.
\newblock \emph{arXiv preprint arXiv:1808.08866}.

\bibitem[{Zhang et~al.(2019)Zhang, Feng, Meng, You, and
  Liu}]{Zhang2019BridgingTG}
Wen Zhang, Y.~Feng, Fandong Meng, Di~You, and Qun Liu. 2019.
\newblock Bridging the gap between training and inference for neural machine
  translation.
\newblock In \emph{ACL}.

\bibitem[{Zoph et~al.(2016)Zoph, Yuret, May, and Knight}]{zoph2016transfer}
Barret Zoph, Deniz Yuret, Jonathan May, and Kevin Knight. 2016.
\newblock \href {http://arxiv.org/abs/1604.02201} {Transfer learning for
  low-resource neural machine translation}.

\end{thebibliography}
\bibliographystyle{acl_natbib}
\newpage
\appendix

\section{Methodology}\label{sec:data_sizes}
% \paragraph{Architecture.}
% \label{subsec:Arch.}
% We use a similar setup as used by \citet{wieting2019beyond}, adapting their fairSeq-based \citep{ott2019fairseq} codebase to our purposes.\footnote{\url{https://github.com/jwieting/beyond-bleu}} Similar to their Transformer architecture we use gated convolutional encoders and decoders \citep{gehring2017convolutional}. We use 4 layers for the encoder and 3 for the decoder, the size of the hidden state is 768 for all layers, and the filter width of the kernels is 3. Additionally, the dimension of the BPE embeddings is set to 768.

% \paragraph{Optimization.}
% We optimize with Nesterov's accelerated gradient method
% \citep{sutskever2013importance} with a learning
% rate of 0.25, a momentum of 0.99, and
% re-normalize gradients to a 0.1 norm \citep{pascanu2012difficulty}.

% \paragraph{Data $\&$ Evaluation.}
% We use a the same data that used by \citet{wieting2019beyond}.
% For training data for cs-en, de-en, and ru-en,
% we use the WMT News Commentary v13\footnote{\url{http://data.statmt.org/wmt18/translation-task}}
% \citep{bojar-etal-2017-findings} for training the models.
% Test sets are the official WMT18 test sets.
% For tr-en, we used for training the
% WMT 2018 parallel data, which consists of the
% SETIMES2 corpus \citep{Tiedemann2012ParallelDT}.
% The validation set is a concatenation of
% newsdev 2016 and 2017 released for WMT18.
Table \ref{tab:data_sizes} present the train, validation and test sizes in all four languages pairs. 
We note that our use of the data is aligned with the license and intended use of the data.

\begin{table}[ht]
%\resizebox{\columnwidth}{!}
{%
\begin{tabular}{@{}llll@{}}
\toprule
Lang. & Train   & Valid & Test                      \\ \midrule
de-en & 284,246 & 6,003 & \multicolumn{1}{c}{2,998} \\
cs-en & 218,384 & 6,004 & 2,983                     \\
ru-em & 235,159 & 5,999 & 3,000                     \\
tr-en & 207,678 & 6,007 & 3,000                     \\ \bottomrule
\end{tabular}%
}
\caption{\label{tab:data_sizes}Number of sentence pairs in the training/validation/test sets for all four languages.}
\end{table}

% We use smoothed BLEU \citep{lin-och-2004-orange} from the Moses implementation to report our results\footnote{\url{https://github.com/jwieting/beyond-bleu/blob/master/multi-bleu.perl}}.
% We note that those results are identical to the results we got from sacrebleu \citep{post-2018-call}\footnote{\url{https://github.com/mjpost/sacrebleu}}.

\section{Entropy of STV and LTV}\label{sec:app_entropy_stv_ltv}
As C19 suggested we can compare the peakiness of the two models by calculating their distributions entropy. Lower entropy indicates a more peaky distribution. We used KL divergence with respect to the uniform distribution in order to normalize the entropy and compare the peakiness of the two models. The STV model starts RL training with mean entropy of 0.300 and finishes with 0.269 while the LTV begins with 0.258 and finishes with 0.264. This indicates that before RL training the LTV model was slightly more peaky than the STV, but after RL training they have similar peakiness.

% \section{STV and LTV in small model}\label{sec:app_stv_ltv_small}
% We also train STV and LTV models with inner dimension of size $d=256$. In this case, as we can see in figure \ref{fig:pro-shift-10-big-small_d256} the LTV is able to shift probability to the first place although STV is doing it more successfully. In addition, within the first one hundred ranks, STV reduces the probability of $88$ of them, whereas LTV only of $77$.  

% % figure 1
% \begin{figure}[ht]
% \centering
% \includegraphics[scale=0.28]{com_p_g_F.png}
% \caption{Comparison of probability shift due to RL training of assign $y_{best}$ for ten first words for both LTV and STV with d=256. in green, you can see the results with BPE of size 1,000, STV. in blue are the results with BPE of size 30,000, LTV. a clear improvement of assigning $y_{best}$ in first place for STV.}
% \label{fig:pro-shift-10-big-small_d256}
% \end{figure}

\section{Loading Bert embedding}\label{sec:app_bert_mle}
We consider two options for initializing Bert embeddings for the MLE training, with and without freezing the embedding layer.
The results were unequivocal, freezing the embedding layers has a very constructive effect on the results (table \ref{tab:table-bert-mle}). We estimate that freezing the embedding layers causes such a vast improvement in performance because it enables us to avoid the catastrophic forgetting of BERT parameters. Therefore, although using BERT embedding is helpful as initialization, by freezing the parameters we allow the model to better utilize BERT's embeddings.

\begin{table}[ht]
\centering
\begin{tabular}{p{2.4cm}|p{0.9cm}|p{0.9cm}|p{0.9cm}|p{0.9cm}}
%{@{}lllll@{}}
\toprule
Model & De-En & Cs-En & Ru-En & Tr-En \\ \midrule
MLE                  & 22.38 & 15.81 & 17.31 & 12.60 \\
MLE+Bert W/o freeze  & 22.99    & 15.32 & 17.57 & 12.65\\
MLE+Bert with freeze &   23.46  & 16.59   & 18.14 & 14.15\\
diff.                & 0.47  & 1.27 & 0.57 & 1.50  \\
 \bottomrule
\end{tabular}
\caption{\label{tab:table-bert-mle} Comparison of MLE models with BERT embedding with and without freezing.}
\end{table}

\section{Number of parameters}\label{sec:app_pram}
In Table \ref{tab:table-bert-pram-comp} we provide a comparison of the number of trainable parameters with and without freezing the embedding layer.
%in two settings, one when we are freezing the embedding layer and the second when we don't freeze.
    
% \begin{table}[ht]
% \begin{tabular}{@{}lllll@{}}
% \toprule
% \# parameters & de-en & cs-en & ru-en & tr-en \\ \midrule
% Freeze        & 11.9M & 11.6M & 11.1M & 11.6M \\
% W/o freeze    & 22.7M & 21.1M & 19.8M & 22.5M \\
% Ratio         & 0.52  & 0.54  & 0.56  & 0.51  \\ \bottomrule
% \end{tabular}

% \caption{\label{tab:table-reg-pram-comp} Comparison if trainable parameters in large bpe setting.}
% \end{table}

% table 4:
\begin{table}[ht]
\centering
\begin{tabular}{@{}lllll@{}}
\toprule
\# parameters & De-En & Cs-En & Ru-En & Tr-En \\ \midrule
Freeze & 30.2M      &  29.3M     &   27.9M    &  29.4M     \\ 
W/o Freeze      &  77.2M     &   76.2M   &   74.8M    &   76.4M    \\ 
Ratio &  0.39     &   0.38    & 0.37      &  0.38  \\  \bottomrule
\end{tabular}
\caption{\label{tab:table-bert-pram-comp} Comparison if trainable parameters.}
\end{table}

\section{Formalizing the intuition behind freezing the embedding layer}\label{sec:app_proof}
Here we want to formalize the intuition behind freezing the embedding layer. We explicitly calculate the gradients of the cross-entropy (CE) loss of the one-hot vector, $y$, and the distribution vector, $\hat y$ of the model $f_{\theta} = h_{\theta_{2}} \circ g_{\theta_{1}}$ output (henceforth, we will discard the parameters notation from $f,g$ and $h$). We will discuss two cases, one when we freeze $\theta_{2}$ and the second when we are not. We note that $\theta_{2} \in \mathbb{R}^{d\times |V_{T}|}$ is the embedding layer where each row, $\rho_i$, is the representation of the $k$’s word in the vocabulary. Moreover, $h: \mathbb{R}^d \rightarrow \mathbb{R}^{|V_{T}|}$ is the function defined by multiplying the output of $g$, denoted by $v \in \mathbb{R}^d$, by $\theta_{2}$, and then taking the soft-max of the output vector, hence $\forall k\in |V_{T}|;  h_k(v) = \frac{exp(\rho_k \cdot v)}{\sum_l exp(\rho_l\cdot v)}$. Therefore assigning to each word some probability, $\hat y_k$, to be the next one in the sentence.

%Similarly, we can define the scalar function $\forall k\in |V_{T}|; f_k: \mathbb{R}^d \rightarrow \mathbb{R}$ that get as an input source sentence, $X$, prefix of the translation, $y_{<i}$, and defined by parameters, $\theta$, and output the probability of the next token to be the $k$’s word in vocabulary. To be more precise, the input of $f_k$ is a vector $p \in \mathbb{R}^d$ that represent $X$ and $y_{<i}$ in the inner dimension of the net. In this terminology, the model chooses the next token to be the one that corresponding to the $f_k$ that upon the input $p$ return the biggest probability. Then we calculating the CE loss between the model distribution and the one hot vector encoding of the gold token, and updating the model parameters according to the gradient of CE.

Now, we want to investigate the update defined by the gradients of the CE loss in the setting when two words, $w_1$ and $w_2$ have the same representation, $\rho_1 = \rho_2$. We consider the case where one of them is the gold token, w.l.g. $w_1$. We note this case by $\Big|^1$.
%In this setting we want to compare the gradients when the model choose $x_1$ to when the model choose $x_2$, corresponding with two input vectors $p_1$ and $p_2$. 
% respectfully, for two input points $p_1$ and $p_2$. 
%reflecting that both of them suit the context and have similar meaning.

We turn to examine the gradient in this setting for both cases. We start by realizing that if all the partial derivatives of the CE loss, $L$, exist then the gradient is the vector of all the partial derivatives meaning, $\nabla_{\theta}L=\begin{pmatrix} \nabla_{\theta_1}L \\ \nabla_{\theta_2}L \end{pmatrix}$ 
and we can separate the calculation into two parts, one with respect to $\theta_1$ and the second with respect to $\theta_2$. 

By definition, in the case where we freeze $\theta_2$ we will keep $\rho_1$ and $\rho_2$ the same. We will now show that in the case when we don't freeze $\theta_2$ the update will be different.
% introduce notation $f(r:\theta, y_{<i}, z)$ where z encoder output (maybe the same to put x, r the input vector for the decoder. for one input we get $x_1$ for another $x_2$.

\begin{lemma}
If $\theta_2$ is not frozen then:
%$\nabla_{\theta_2}L\Big|_{p_1} \neq \nabla_{\theta_2}L\Big|_{p_2}$. specifically, 
Updates are differe: $\Delta\rho_1!=\Delta\rho_2$.
%consequentially, we  
% we update $\rho_1$ differently from $\rho_2$ upon our choose. 
\end{lemma}

\begin{proof}
We start by noticing that multiplying $v$ by $\theta_2$ is a linear transformation so for points $p_1$ and $p_2$ we will get the same derivative as $\rho_1 = \rho_2$, moreover by taking the soft-max of those identical outputs we will get the same outputs. Hence, we get that $\forall i\in [d];  \frac{\partial \hat y_1}{\partial v_i} = \frac{\partial \hat y_2}{\partial v_i}$, similarly $\frac{\partial \hat y_1}{\partial \rho_{1i}} = \frac{\partial \hat y_2}{\partial  \rho_{2i}}$. 

We continue by calculating the derivative of the CE. 
The CE loss is defined by:
\begin{equation}
    L(y, \hat y) = \sum_i y_i log(\hat y_i)
\end{equation}
The derivative is: $\frac{\partial L}{\partial \hat y_i}= \sum_i y_i \frac{1}{\hat y_i}$ we notice that  $y$ is a one hot vector i.e., $y_1=1$ and $\forall i\in [2,|V_{T}|]; y_i = 0$. Therefore, the derivative will be different from $i=1$ to all other $i$'s. Specifically, $\forall i\in [d]; \frac{\partial L}{\partial \rho_{1i}} \neq \frac{\partial L}{\partial \rho_{2i}}$ .
Putting it all together we get:
\begin{equation}
    \frac{\partial L}{\partial \rho_{1i}} = \frac{\partial L}{\partial \hat y_1} \cdot \frac{\partial \hat y_1}{\partial \rho_{1i}} \neq \frac{\partial L}{\partial \hat y_2} \cdot \frac{\partial \hat y_2}{\partial \rho_{2i}} = \frac{\partial L}{\partial \rho_{2i}}
\end{equation}
Proving that $\rho_1$ and $\rho_2$ updates are different.

\end{proof}

\begin{lemma}
For both cases, the update of $\theta_1$ is symmetric to the gold being $w_1$ or $w_2$. 
%$\nabla_{\theta_1}L = \nabla_{\theta_1}L$ meaning, the update of $\theta_1$ we 
$\nabla_{\theta_2}L\Big|^1 = \nabla_{\theta_2}L\Big|^2$.
%$\nabla_{\theta_1}f$ is the same regardless of model choosing $x_1$ or $x_2$.
\end{lemma}

\begin{proof}
Given a parameter $\lambda \in \theta_1$, we inspect the derivative of $L$ with respect to $\lambda$. We use here Einstein summation notation.
\begin{equation}
    \frac{\partial L}{\partial \lambda}\Big|^1 = \frac{\partial L}{\partial v_i} \cdot \frac{\partial v_i}{\partial \lambda}\Big|^1 = \frac{\partial L}{\partial v_i} \cdot \frac{\partial v_i}{\partial \lambda}\Big|^2 = \frac{\partial L}{\partial \lambda}\Big|^2
\end{equation}
We deduce $\frac{\partial v_i}{\partial \lambda}\Big|^1 = \frac{\partial v_i}{\partial \lambda}\Big|^2$, as the derivative of $v_i$ is independent of the question which word is the gold.
In order to justify the second equality we used, we will write the derivative of $L$ with respect to $v_i$.
\begin{equation}
    \frac{\partial L}{\partial v_i}\Big|^1 = \frac{\partial L}{\partial \hat y_k} \cdot \frac{\partial \hat y_k}{\partial v_i}
\end{equation}
Clearly, we only need to check the elements that change by switching the gold from being $w_1$ to $w_2$ or vice versa. Therefor all the second terms that multiply by $\frac{\partial L}{\partial \hat y_k}$ for $k\in [3,|V_{T}|]$ didn’t change. We already proved that $\forall i\in [d];  \frac{\partial \hat y_1}{\partial v_i} = \frac{\partial \hat y_2}{\partial v_i}$ 
%what translated here to $\frac{\partial \hat y_1}{\partial \rho_{1j}} \cdot \frac{\partial \rho_{1j}}{\partial v_i} = \frac{\partial \hat y_2}{\partial \rho_{2j}} \cdot \frac{\partial \rho_{2j}}{\partial v_i}$. 
Finally, because we switch the gold, $\frac{\partial L}{\partial \hat y_1}$ and $\frac{\partial L}{\partial \hat y_2}$ indeed switch there values but both of them are multiply by the same values as $\frac{\partial \hat y_1}{\partial v_i} = \frac{\partial \hat y_2}{\partial v_i}$. Overall, the derivative is unchanged.
\end{proof}

To conclude, in the motivational setting we discussed, when we freeze $\theta_2$ we keep semantically close vectors unchanged while if we don't freeze $\theta_2$ we enable them to change. As consequence, in further steps, this change will affect on $\theta_1$ also. In a similar manner, as long as the representation is similar, all layers but the penultimate would update both words similarly.

% We can calculate $\nabla_{\theta}f=
% \begin{pmatrix}
% \nabla_{\theta_1}f \\ \nabla_{\theta_2}f 
% \end{pmatrix}$ as all partial derivatives of $f$ exist.
% because all partial derivatives of a function exist  

% Denote the function that the network computes with $f_{\theta}$.
% $f_{\theta}$ can be written as $h_{\theta_{2}} \circ g_{\theta_{1}}$,
% where $\theta = \left( \theta_{1}, \theta_{2} \right)$, $g$ maps the input -- source sentence $X$
% and model translation prefix $y'_{<i}$ -- into $\mathbb{R}^d$, and $h$ maps $g$’s output into $\mathbb{R}^{|V_{T}|}$.

\section{Semantic scores for the second method}\label{sec:app_bert_sim}

% % table 1:
% \begin{table}[ht]
% \centering
% \begin{tabular}{ @{}lrrrr@{}  }
%  \toprule
%  Model & de-en & cs-en &ru-en& tr-en\\
%  \midrule
%  LTV  & 80.09    & 74.48 & 78.17 & 69.32\\
%  LTV+RL&   80.84  & 75.45   & 79.42 & 69.93\\
%  Diff. & 0.75 & 0.97 &  1.25 & 0.61\\
%  \midrule
%  STV  & 98.95    & 98.64 & 98.97 & 98.45\\
%  STV+RL&   99.07  & 98.68   & 99.01 & 98.55\\
%  Diff. & 0.12 & 0.04 &  0.04 & 0.10\\

%  \bottomrule
% \end{tabular}
% \caption{\label{tab:table-small-big-results} SIM scores}
% \end{table}

% table 1:
% \begin{table}[ht]
% \centering
% \begin{tabular}{ @{}lrrrr@{}  }
%  \toprule
%  Model & de-en & cs-en &ru-en& tr-en\\
%  \midrule
%  LTV  & 75.87    & 70.55 & 73.57 & 65.19\\
%  LTV+RL&   76.89  & 71.50   & 74.96 & 65.82\\
%  Diff. & 1.02 & 0.95 &  1.39 & 0.63\\
%  \midrule
%  STV  & 93.54    & 92.92 & 92.43 & 91.50\\
%  STV+RL&   94.50  & 93.57   & 93.23 & 93.13\\
%  Diff. & 0.96 & 0.65 &  0.80 & 1.63\\

%  \bottomrule
% \end{tabular}
% \caption{\label{tab:table-small-big-results} smaile scores}
% \end{table}

%We present the SIM results for the second method. SIM is a measure of semantic similarity that assigns partial credit to semantically correct but lexically different translations \citep{wieting2019beyond}.
Our method of freezing informative initialization of the embedding layer aims to generalize across different but semantically close actions. In order to test the ability of our model to generalize we used SIM. SIM is a measure of semantic similarity that assigns partial credit to semantically correct but lexically different translations \citep{wieting2019beyond}.
%As our second method aims to generalize across different but semantically close actions, this score function can reflect this ability.
Table \ref{tab:table-BERT-sim-results} shows our model results and exhibits similar trends to the BLEU scours. Here we see even greater gains for cs-en and ru-en languages pairs. Those results may indicate that the model was able to predict tokens that are semantically close to the gold token.

\begin{table}[ht]
\large
\resizebox{\columnwidth}{!}{%
\begin{tabular}{@{}lllll@{}}
\toprule
MODEL           & De-En & Cs-En & Ru-En & Tr-En \\ \midrule
MLE             & 70.03 & 63.29 & 66.17 & 59.68 \\
+RL             & 71.17 & 63.29 & 66.17 & 59.99 \\
% \hline
\midrule

% \hline
% \hline
+RL+FREEZE      & 71.03 & 64.29 & 66.66 & 60.52 \\
% \midrule
+BERT          & 71.56 & 64.26 & 66.70 & 61.75 \\
+BERT+RL       & 72.44 & 65.80 & \textbf{67.94} & \textbf{63.59} \\
+BERT+RL+FREEZE & \textbf{72.81} & \textbf{66.44} & 67.66 & \textbf{63.59} \\ \bottomrule
\end{tabular}%
}\caption{\label{tab:table-BERT-sim-results}SIM scores on translating four languages to English.}
\end{table}

% \begin{table}
% \large
% \resizebox{\columnwidth}{!}{%
% \begin{tabular}{@{}lllll@{}}
% \toprule
% MODEL           & de-en & cs-en & ru-en & tr-en \\ \midrule
% MLE             & 66.84 & 60.20 & 62.97 & 56.25 \\
% +RL             & 68.37 & 60.20 & 62.97 & 56.82 \\
% % \hline
% \midrule

% % \hline
% % \hline
% +RL+FREEZE      & 68.24 & 61.44 & 63.50 & 57.38 \\
% % \midrule
% +BERT           & 68.57 & 61.46 & 63.42 & 58.28 \\
% +BERT+RL       & 69.54 & 64.80 & \textbf{75.05} & 60.28 \\
% +BERT+RL+FREEZE & \textbf{69.92} & \textbf{72.78} & 64.40 & \textbf{60.28} \\ \bottomrule
% \end{tabular}%
% }\caption{\label{tab:table-BERT-results}smile}
% \end{table}

\section{Human Evaluation Information} \label{sec:human_eval_info}
We recruited the service of two professional translators via translations providers.
\subsection{Human Evaluation Instructions}
You will be shown:

\begin{enumerate}
    \item An English segment of text;
    \item Corresponding translation into English.
\end{enumerate}

There are three parts to each annotation:
\begin{enumerate}
    \item Read the English segment;
    \item Read the translation and compare its meaning to the meaning of the original English segment;
    \item Give a score between 0-100 describing how close the meaning of the translation is to the meaning of the original English segment.
\end{enumerate}

% \bibliographystyle{acl_natbib}

%This is an appendix.

\end{document}